\documentclass[preprint,12pt,authoryear]{elsarticle}

\usepackage{geometry}
\usepackage{amssymb}
\usepackage[utf8]{inputenc} 
\usepackage[T1]{fontenc}    
\usepackage{hyperref}       
\usepackage{url}            
\usepackage{booktabs}       
\usepackage{amsfonts}       
\usepackage{nicefrac}       
\usepackage{microtype}      
\usepackage{xcolor}         
\usepackage{epsfig}
\usepackage{graphicx}
\usepackage{amsmath}
\usepackage{graphicx}
\usepackage{amssymb}
\usepackage{booktabs}
\usepackage{graphicx}
\usepackage{float}
\usepackage{stfloats}
\usepackage{epstopdf}
\usepackage{algorithm,algorithmic}
\usepackage{multirow}
\usepackage{bbm}
\usepackage{framed}
\usepackage{amsthm}
\usepackage{adjustbox}
\usepackage{color}
\usepackage{lineno}
\usepackage{url}
\usepackage{cleveref}
\usepackage{diagbox}

\usepackage{colortbl}
\usepackage{arydshln}
\usepackage{multirow}
\usepackage{multicol}
\usepackage{makecell}
\usepackage{upgreek}
\usepackage{gensymb}
\usepackage{lineno}
\usepackage{makecell}

\begin{document}

\begin{frontmatter}




\title{Large-scale flood modeling and forecasting with FloodCast}

\author[label1]{Qingsong Xu}
\author[label2]{Yilei Shi}
\author[label1,label3]{Jonathan Bamber}
\author[label4]{Chaojun Ouyang}
\author[label1]{Xiao Xiang Zhu\corref{cor1}}
\affiliation[label1]{organization={Data Science in Earth Observation, Technical University of Munich},
            state={Munich 80333},
            country={Germany}}
\affiliation[label2]{organization={Technical University of Munich},
            state={Munich 80333},
            country={Germany}}
\affiliation[label3]{organization={School of Geographical Sciences, University of Bristol},
            state={Bristol BS8 1SS},
            country={UK}}
\affiliation[label4]{organization={Institute of Mountain Hazards and Environment, Chinese Academy of Sciences}, state={Chengdu 610041},
            country={China}}
\cortext[cor1]{Corresponding author.}
\fntext[]{E-mail addresses: qingsong.xu@tum.de (Qingsong Xu), yilei.shi@tum.de (Yilei Shi), J.Bamber@bristol.ac.uk (Jonathan Bamber), cjouyang@imde.ac.cn (Chaojun Ouyang), xiaoxiang.zhu@tum.de (Xiao Xiang Zhu)}



\begin{abstract}
Large-scale hydrodynamic models generally rely on fixed-resolution spatial grids and model parameters as well as incurring a high computational cost. This limits their ability to accurately forecast flood crests and issue time-critical hazard warnings. In this work, we build a fast, stable, accurate, resolution-invariant, and geometry-adaptative flood modeling and forecasting framework that can perform at large scales, namely FloodCast. The framework comprises two main modules: multi-satellite observation and hydrodynamic modeling. In the multi-satellite observation module, a real-time unsupervised change detection method and a rainfall processing and analysis tool are proposed to harness the full potential of multi-satellite observations in large-scale flood prediction. In the hydrodynamic modeling module, a geometry-adaptive physics-informed neural solver (GeoPINS) is introduced, benefiting from the absence of a requirement for training data in physics-informed neural networks (PINNs) and featuring a fast, accurate, and resolution-invariant architecture with Fourier neural operators.
To adapt to complex river geometries, we reformulate PINNs in a geometry-adaptive space. 
GeoPINS demonstrates impressive performance on popular partial differential equations across regular and irregular domains. Building upon GeoPINS, we propose a sequence-to-sequence GeoPINS model to handle long-term temporal series and extensive spatial domains in large-scale flood modeling. This model employs sequence-to-sequence learning and hard-encoding of boundary conditions. Next, we establish a benchmark dataset in the 2022 Pakistan flood using a widely accepted finite difference numerical solution to assess various flood prediction methods. Finally, we validate the model in three dimensions - flood inundation range, depth, and transferability of spatiotemporal downscaling - utilizing SAR-based flood data, traditional hydrodynamic benchmarks, and concurrent optical remote sensing images. Traditional hydrodynamics and sequence-to-sequence GeoPINS exhibit exceptional agreement during high water levels, while comparative assessments with SAR-based flood depth data show that sequence-to-sequence GeoPINS outperforms traditional hydrodynamics, with smaller prediction errors.
The experimental results for the 2022 Pakistan flood demonstrate that the proposed method enables high-precision, large-scale flood modeling with an average MAPE of 14.93\% and an average MAE of 0.0610m for 14-day water depth predictions while facilitating real-time flood hazard forecasting using reliable precipitation data. FloodCast is publicly available at \url{https://github.com/HydroPML/FloodCast}.
\end{abstract} 

\end{frontmatter}


\section{Introduction}
Flooding is a frequent and widespread natural hazard, annually causing substantial human and property losses globally. In 2022 alone, over 57 million people were affected, resulting in 8,000 fatalities and \$45 billion in damages~\citep{fraehr2023supercharging}. This underscores the urgent need for effective flood warning and control systems. Flood forecasting is a crucial tool for delivering timely hazard information to government decision-makers, practitioners, and at-risk residents, playing a vital role in reducing flood-related risks~\citep{carsell2004quantifying}. Considerable efforts have been invested in developing forecasting systems for various flood types, including fluvial, coastal~\citep{muste2022flood}, and urban floods~\citep{henonin2013real}. Regardless of the specific type of flood under consideration, a comprehensive flood forecasting system typically comprises a minimum of two key components: the parameterization and calibration of flood-related variables or drivers, such as precipitation, initial conditions, and a hydrological or hydraulic model to efficiently simulate the flooding processes along the river networks and in the floodplains. 

Parameterization and calibration of flood-related variables have conventionally relied on surveyed or measurement data (e.g., topographic data, initial water level height). However, measurement data is not available in most parts of the world. Furthermore, ground surveys are very expensive and time-consuming in large-scale regions. Therefore, model parameters cannot be well constrained which generally leads to the problem of equifinality and induces large uncertainty in the calibration~\citep{jiang2019simultaneous}. This impedes lagle-scale hydrodynamic modeling and operational flood forecasting for poorly monitored basins.

In recent decades, advances in remote sensing observations have played an increasingly prominent role in hydrologic and hydraulic fields, and have promoted the estimates of key variables related to flood forecast (e.g., initial water extents and depths~\citep{cohen2018estimating}, satellite-based precipitation~\citep{pradhan2022review}, Manning’s roughness~\citep{schumann2009progress}, and river width, etc.). 
Using satellite observations can help better constrain model parameters and reduce calibration uncertainty because satellite observations have higher spatial coverage and contain information on land-surface processes that cannot be inferred from discharge~\citep{wanders2014benefits}. Satellite-based technologies encompass a wide array of sensors, both active and passive, operating across various electromagnetic spectrum regions, including visible, thermal, and microwave domains. 
Advanced sensors such as Synthetic Aperture Radars (SAR), Satellite-based rainfall, and gravitational measurements are emerging as potential game-changers for extreme event forecasting and monitoring. Notably, SAR has the advantage of monitoring large-scale surface water extents in nearly all weather conditions~\citep{le2022multiscale}, such as during flood events when optical satellite visibility is hindered by cloud cover. 
It effectively maps water extents and delineates inundated regions by utilizing the specular reflection of microwave frequencies over smooth water surfaces~\citep{gupta2022pakistan}.
Many existing algorithms have been used in the extraction of flood inundation from SAR imagery, such as threshold determination~\citep{chini2017hierarchical, thonfeld2016robust}, change detection~\citep{le2022multiscale, xu2023ucdformer}, supervised or unsupervised machine learning (ML)~\citep{jiang2021rapid, bentivoglio2022deep,xu2023disasternets}. Furthermore, satellite-based approaches provide the best alternative for estimating and forecasting precipitation in areas with low or no rain gauge density. Several gridded precipitation datasets are developed from satellite data globally, such as 
the Tropical Rainfall Measuring Mission (TRMM)~\citep{huffman2007trmm}, precipitation nowcasting using neural networks and radar observations~\citep{zhang2023skilful}, and the Global Precipitation Measurement Integrated Multi-satellite Retrievals (GPM-IMERG)~\citep{hou2014global}.
However, the majority of studies primarily focus on identifying hydrodynamic resistance, such as  Manning's roughness coefficient calculated from land use and land cover~\citep{schumann2009progress, ming2020real}. This emphasis often neglects a thorough exploration of the pivotal role that satellite observation data can play in advancing flood dynamics prediction. Additionally, many SAR-based flood mapping approaches necessitate data annotation and manual parameter adjustments, such as threshold values, which hinder the swift automated mapping of extensive and intricate flood inundation areas. Furthermore, the limited spatiotemporal resolution of satellite-based rainfall data presents challenges in real-time processing for high spatiotemporal resolution flood simulations.

Prediction of flooding processes induced by rainfall is another essential component in a flood forecasting system, which may involve the use of a wide variety of hydraulic or hydrodynamic models~\citep{teng2017flood}. Accurate flood inundation predictions are traditionally provided
by running two-dimensional hydrodynamic models that solve shallow water equations (SWEs)~\citep{wang2022rapid}. However,  traditional hydrodynamic methods (e.g. finite difference (FD)~\citep{van2010lisflood, de2013applicability}) are highly dependent on the resolution: coarse grids are fast but low accurate; fine grids are accurate but slow. For example, a global flood solver based on FD has a resolution of up to 1 km, but this requires a dramatic increase in computational resources~\citep{zhou2021toward}. While extensive efforts have been made to develop efficient numerical schemes to represent a more dynamic river system in large-scale models, such as high-performance computing~\citep{schubert2022framework}, parallelization, graphics card processing~\citep{ming2020real}, the alternative is to address the scalability challenge of hydrodynamic methods in regions of interest using ML or physics-aware ML hybrid approaches.

A ``good'' hydrodynamic solver for large-scale simulations should satisfy the following basic conditions: $i$) Fast and accurate. Stable and accurate approximation schemes can be obtained with minimal computational overhead; $ii$) Resolution-invariant. The solver can efficiently represent dynamic processes at different spatial and temporal resolutions; $iii$) Geometry-adaptive. The solver can be adapted to different rivers, where the geometry is often complex and irregular. Motivated by this, there has been a recent interest in developing ML approaches to find solutions for SWEs or partial differential equations (PDEs) (and/or work in tandem with numerical solutions). Recently, deep neural networks (DNNs) have been demonstrated to be promising for solving fluid dynamics systems~\citep{he2023deep,xu2023physics}.  A recent work involves physics-informed neural networks (PINNs)~\citep{raissi2019physics, feng2023physics} by approximating solution functions and neural operators~\citep{anandkumar2020neural,li2020fourier,xu2023large} by learning solution operators.
However, the application of ML or physics-aware ML to
large-scale modeling is still limited because (a) large-scale hydrodynamic models have a complex terrain surface, land use, and land cover and depend on a large number of model parameters; and (b) to date, there is no consistent ML method to integrate both remote sensing data and in-situ measurements. Furthermore, PINNs are mainly defined in a pointwise way. Despite their noticeable empirical success, they may fail to solve challenging PDEs when the solution exhibits high-frequency components~\citep{krishnapriyan2021characterizing, fuks2020limitations}.    Neural operator methods, such as Fourier neural operator (FNO)~\citep{li2020fourier}, treat the mapping from initial conditions to solutions as an input–output mapping that can be learned by supervision. FNO is a resolution-invariant (zero-shot super-resolution) operator in the spatial and temporal domains. Moreover, due to the fast Fourier transformation (FFT), FNO is a fast architecture that allows to obtain accurate results for SWEs. However, FNO carries out supervised learning on a given dataset, which is infeasible for flood simulation. 
Thus, these solvers could not fulfill the properties of a "good" hydrodynamic solver.

Bearing these concerns in mind,  we build a fast, stable, accurate, resolution-invariant, and geometry-adaptative flood modeling and forecasting framework that can perform at large scales, namely FloodCast.  The framework includes the multi-satellite observation and hydrodynamic modeling modules. Combining these two modules, the system can produce reliable, real-time temporal‐spatial varying flood inundation depths and extents across a predefined simulation domain. Specifically, to harness the full potential of multi-satellite observations in large-scale flood prediction, we have developed a dedicated module with three key roles in flood modeling and forecasting.
First, the hydrodynamics model is driven by reliable model parameters from multi-satellite observations to predict the large-scale flooding processes across the simulation domain.  A real-time end-to-end rainfall processing and analysis tool is designed from the satellite-based precipitation products to obtain real-time precipitation at any resolution in the study area. In addition, other relevant datasets including land cover are also required to estimate model parameters.
Second, a high temporal and spatial water inundations and depths estimated module is presented based on a real-time unsupervised change detection (UCD) method and Sentinel-1 SAR images. The UCD algorithm exploits available SAR images in combination with historical Landsat and other auxiliary data sources hosted on the Google Earth Engine (GEE), to rapidly map time-series water inundation. Furthermore,  the SAR-based flood extent and terrain data are utilized to generate water inundation depths for starting the hydrodynamic model. Third, the high temporal, high spatial water inundations and depths from multi-satellite observations can be employed to calibrate and validate the hydrodynamic model. 

For the hydrodynamic modeling in FloodCast, aiming to construct a "good" hydrodynamic solver for flood simulations, we propose a geometry-adaptive physics-informed neural solver (GeoPINS), that overcomes the shortcomings of both PINNs and neural operators. Compared to PINNs, GeoPINS has a  better spatial-temporal representation space,
so, GeoPINS converges faster and more accurately. It achieves geometry adaptation and resolution invariance when solving PDEs. Building upon GeoPINS, we propose a sequence-to-sequence GeoPINS model to handle long-term temporal series and extensive spatial domains in large-scale flood modeling. This involves employing sequence-to-sequence learning for the temporal dimension and a hard-encoding of boundary conditions (e.g., river inflow boundary). 

We utilize the FloodCast framework to predict and analyze the severe floods that occurred in Pakistan between August and September 2022, attributed to heavy rainfall. Specifically, we initially employ the proposed high-precision, real-time UCD to derive a time series of SAR-based flood inundation extents spanning from August 6 to September 11, 2022. A pronounced change in flood coverage is observed between August 18 and August 30, marked by significant increases. Subsequently, we analyze the spatiotemporal rainfall intensity in Pakistan during the same period (August 18 to August 30) to determine the necessity of utilizing the FloodCast model. We identify that the rainfall during this timeframe surpassed predefined extreme rainfall thresholds, underscoring the critical need for accurate flood prediction with high spatiotemporal resolution. Then, to rapidly simulate the study area, coarse spatiotemporal resolutions (480m $\times$ 480m spatial resolution and 300s temporal resolution) over a 14-day period, from August 18th to August 31st are employed.  Additionally, we establish a benchmark dataset in the Pakistan study region using a widely accepted FD numerical solution technique to assess various flood prediction methods. We validate the model in three dimensions - flood inundation range, depth, and transferability of spatiotemporal downscaling - using SAR-based flood data, traditional hydrodynamic benchmarks, and concurrent optical remote sensing images. Our analysis reveals that deterministic hydrodynamics-based flood extents align better with SAR-based results in peripheral regions, while sequence-to-sequence GeoPINS-based flood extents excel in accuracy within inundated areas, especially in detail. Both methods exhibit exceptional agreement during high water levels. Additionally, we conduct a comprehensive assessment of the accuracy and effectiveness of our proposed sequence-to-sequence GeoPINS in predicting water depth across large spatial and temporal scales. Comparative evaluations with SAR-based flood depth data indicate that sequence-to-sequence GeoPINS outperforms traditional hydrodynamics, displaying smaller prediction errors. Furthermore, the proposed sequence-to-sequence GeoPINS demonstrates significantly enhanced accuracy in predicting spatially varying water depths, especially in cases of elevated water levels.

Our contributions can be summarized as follows:
\begin{itemize}
	\item We have developed the FloodCast framework for large-scale flood modeling and forecasting. This framework includes multi-satellite observation and hydrodynamic modeling modules. In the multi-satellite observation module, a real-time unsupervised change detection method and a rainfall processing and analysis tool are proposed to harness the full potential of multi-satellite observations in large-scale flood prediction.
	\item We propose the GeoPINS, that combines the advantages of leveraging PINNs to overcome the need for training data and leveraging neural operators. Building upon GeoPINS, we propose a sequence-to-sequence GeoPINS model to handle long-term temporal series and extensive spatial domains in large-scale flood modeling. 
	\item We establish a benchmark dataset using a widely accepted finite difference numerical solution technique to assess various flood prediction methods. 
	\item We construct a novel flood prediction and analysis process on the 2022 Pakistan flood, using SAR-based flood data, traditional hydrodynamic benchmarks, and concurrent optical remote sensing images. Our experimental results demonstrate that the proposed method consistently maintains high-precision solutions for large-scale flood dynamics. Furthermore, we employ a model trained on a coarse grid to infer directly on a refined spatiotemporal grid, effectively addressing the scalability challenges associated with traditional hydrodynamics. The effectiveness of zero-shot super-resolution experiments holds the potential for real-time flood prediction on high spatiotemporal resolution grids.
\end{itemize}

\section{Related Work}
As represented by the hydrodynamics module (GeoPINS) in FloodCast, a popular approach to combine machine learning and physical knowledge is to train different NNs by minimizing the violation of physical laws, when the governing PDEs (such as shallow water equations)  are known.  The effectiveness of the fully connected neural networks (FCNN) based PDE solution algorithm has been demonstrated on a number of  PDEs~\citep{lu2021deepxde, jin2021nsfnets, moseley2020solving, yang2020physics, cai2022physics, xu2023physics}. Despite some success, there have also been issues observed with the FCNN formulation (e.g. spectral bias)~\citep{wang2022and, krishnapriyan2021characterizing}. To enable efficient learning of large-scale spatiotemporal solution fields, the convolutional neural networks (CNNs) and FNOs~\citep{kim2019deep, wandel2021spline, li2021physics} have attracted increasing attention in the PINNs' community. Among these neural networks, traditional numerical approaches (e.g. FD~\citep{wandel2020learning}, finite volume (FV)~\citep{gao2022physics}, and pseudo-spectral numerical methods~\citep{li2021physics,rosofsky2023applications}) are often used for space-time gradient solving in the physics-informed loss. However, all the existing studies on physics-informed CNNs and FNOs are only able to deal with problems defined on regular (rectangular) and small-scale domains. In order to learn solutions of PDEs on irregular domains by physics-informed NNs, PhyGeoNet~\citep{gao2021phygeonet} is proposed to enable data-free learning for PDEs with irregular geometries.  Geo-FNO~\citep{li2022fourier} is designed by learning to deform the input (physical) domain. Domain agnostic FNO~\citep{liu2023domain} is proposed by incorporating a smoothed characteristic function in the integral layer architecture of FNOs~\citep{li2020fourier}. 
Nevertheless, these methods cannot deal with dynamic systems at large scales and are unstable on complex geometries and noisy data. 
Further, PINN-based approaches also exploit graph representations in terms of graph neural networks~\citep{harsch2021direct, gao2022physics}, to solve PDEs on irregular domains. Unfortunately, graph neural networks cannot make use of the highly efficient implementations for convolutional operations on grids and thus usually come with higher computational complexity per node compared to grid-based CNNs.
\section{Methodology}

\begin{figure}[!htp]
	\centering
	{\includegraphics[width = 1.05\textwidth]{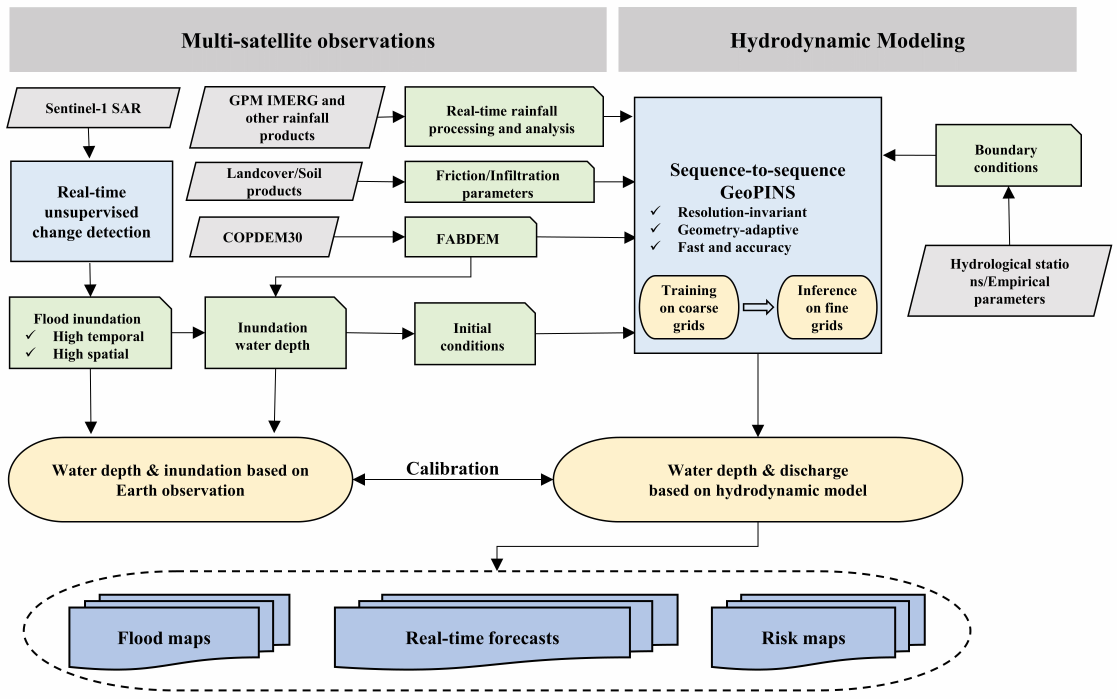}}
	\vspace{-2mm}
	\caption{Overall of flood modeling and forecasting (FloodCast) system.}
	\label{fig:0}
	\vspace{-4mm}
\end{figure}

The structure of the proposed flood modeling system (FloodCast) is illustrated in Fig.~\ref{fig:0}. The framework includes the multi-satellite observation module and hydrodynamic modeling. Integrating these two modules enables the system to consistently produce real-time spatiotemporal variations in flood depths across a predefined simulation domain, such as a catchment area or an entire country. This data can then undergo additional processing to generate inundation maps and other critical flood-related information essential for issuing flood warnings. The results can be also used to support flood risk analysis by superimposing the relevant vulnerability and exposure data.

\subsection{Multi-satellite Observations}
The multi-satellite observation module plays three roles in flood modeling and forecasting. First, the hydrodynamics model is driven by reliable model parameters from multi-satellite observations to predict the large-scale flooding processes across the simulation domain. This includes a high-resolution forest and buildings removed Copernicus digital elevation model (FABDEM)~\citep{hawker202230} from COPDEM30, as well as  real-time precipitation at any resolution from sources like GPM-IMERG with a half-hourly
temporal resolution and $0.1^{\circ} \times 0.1^{\circ}$ spatial resolution~\citep{hou2014global},  or other rainfal products, using a real-time end-to-end rainfall processing and analysis tool.
In addition, other relevant datasets including land cover and soil properties are also required to estimate model parameters.
Second, a high temporal and spatial water inundations and depths estimated module is presented based on a novel real-time UCD method and earth observation data. Specifically,  the proposed UCD algorithm exploits available Sentinel-1 SAR images in combination with historical Landsat and other auxiliary data sources hosted on the GEE, to rapidly map time-series water inundation. Furthermore,  the SAR-based flood extent and FABDEM are utilized to generate water inundation depths based on an automated Floodwater Depth Estimation Tool (FwDET)~\citep{cohen2018estimating}.  
The estimated water inundations and depths provide initial conditions for starting the hydrodynamic model. 
Third, the high temporal, high spatial water inundations and depths from multi-satellite observations can be used to calibrate and validate the hydrodynamic model.

\subsubsection{Real-time unsupervised change detection}
\begin{figure}[!htp]
	\centering
	{\includegraphics[width = 1.05\textwidth]{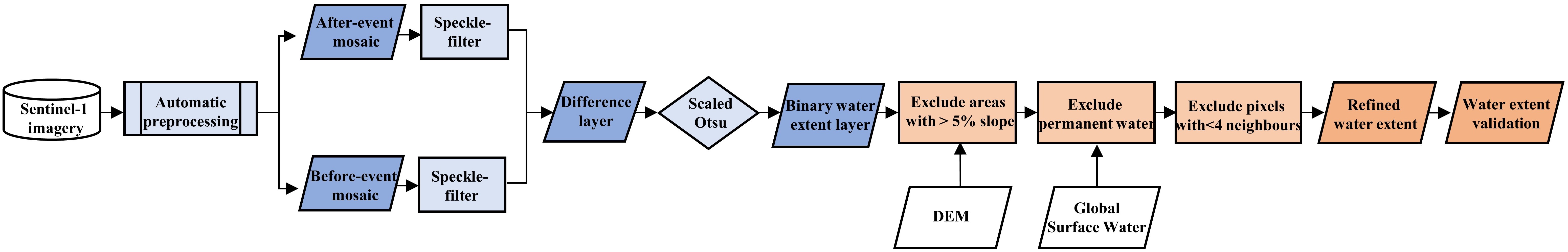}}
	\vspace{-2mm}
	\caption{Overall of unsupervised change detection.}
	\label{fig:12}
	\vspace{-4mm}
\end{figure}

In SAR imagery, open calm water is associated with low backscatter values since there is a strong specular reflection on the water surface. As a result, only a comparatively small proportion of backscatter energy returns to the sensor. In contrast, the backscatter of the surrounding areas normally has relatively high backscatter due to rough surfaces. Based on this characteristic and the emergence of cloud computing platforms (GEE), we propose a fully automatic and real-time change-detection-based algorithm, which is shown in Fig.~\ref{fig:12}. The proposed algorithm tackles issues not previously addressed~\citep{chini2017hierarchical,le2022multiscale} that relate to the objective of real-time and automatic processing of large collections of images. 
The procedures of the proposed UCD method is divided into four main steps. 

The first step is the data preprocessing. The Sentinel-1 time series is preprocessed by several steps before applying the UCD. First, Sentinel-1 Ground Range Detected products are obtained by a standard generic workflow~\citep{filipponi2019sentinel} in GEE, including apply-orbit-file, border noise removal, thermal noise removal, radiometric calibration,  terrain correction, and conversion to decibels (dB). Then, a smoothing filter is applied to reduce the inherent speckle effect of radar imagery. 

The second step is the detection of the water area, which can be performed using threshold-based UCD.  Specifically, the difference layer is computed  following the popular technique of change vector analysis~\citep{saha2019unsupervised,xu2023ucdformer},
\begin{equation}
d(V_{a}, V_{b})=\frac{1}{C}\sum_{i=1}^{C} \left\|V_{a}-V_{b}\right\|_2^2,
\end{equation} 
where $V_{a}$ and $V_{b}$ represent the after-flood mosaic and the before-flood mosaic, respectively. $C$ is the channel of the imagery. $d(V_{a}, V_{b})$ denotes the difference layer. Then, the difference layer is normalized to $[0,1]$ by the min-max scaling. Then, $d(V_{a}, V_{b})$ are divided into two groups using a threshold value ($\mathcal{T}$) to obtain two sets $\Omega_w$ for changed pixels and $\Omega_{nw}$ for unchanged pixels. Any suitable thresholding technique can be utilized, such as Otsu~\citep{thonfeld2016robust}. Otsu’s thresholding is a popular method to determine the decision boundary between changed and unchanged pixels. However, Otsu’s global thresholding lacks adaptability and flexibility for different research areas. To address this problem, we have proposed a scaled Otsu’s thresholding in our work. Specifically, 
\begin{equation}
\mathcal{T} = K*\mathcal{T}_{otsu},
\end{equation} 
where $k$ denotes the adaptive scale factor for different regions. 
$k$ is set to 1.65 by default in our experiments.
Then, the set of pixels $d(V_{a}, V_{b})$ is classified into $\Omega_w$ and $\Omega_{nw}$ according to the following rule,
\begin{equation}
d(V_{a}, V_{b}) \in \begin{cases}\Omega_w, & \text { if } d(V_{a}, V_{b}) \geq \mathcal{T} \\ \Omega_{nw} & \text { otherwise. }\end{cases}
\end{equation}

The third step is the water extent improvement and refinement.  In this step, we consider additional information. For example,  a digital elevation model is utilized to remove areas with over 5 \% slope. The JRC Global Surface Water dataset is used to mask out all areas covered by water for more than 10 months per year. The connectivity of the flood pixels is assessed to eliminate those connected to four or fewer neighbors.

The fourth step is the water map validation. This step is performed only when official flood maps or field survey maps are available. Finally,  using the UCD method, the time-series flood inundation maps  
can be obtained in real-time for the hydrodynamic model. 
\subsubsection{Real-time Rainfall Processing and Analysis}
Fast access to reliable rainfall data can effectively improve the skills of flood forecasting. 
To this end, a real-time end-to-end rainfall processing and analysis tool is designed to swiftly obtain rainfall data across diverse spatial resolutions within the designated study area. The tool comprises automatic rainfall crawling, rainfall refinement processing, and rainfall spatio-temporal analysis. Specifically, first, we perform automated retrieval and download of relevant rainfall data from suitable precipitation products. GPM-IMERG with a half-hourly
temporal resolution and $0.1^{\circ} \times 0.1^{\circ}$ spatial resolution is employed in FloodCast. Then, in the stage of rainfall refinement processing, we employ bilinear interpolation to resample the data, aligning with the spatial resolutions of topographic data in the study area. Additionally, to maintain thorough alignment with the research area's scope, the rainfall data is refined and extracted through the utilization of a spatial mask tool. Finally, the temporal and spatial analysis of the rainfall data is conducted to initially assess the potential for flooding within the study area. More precisely, for the study area, if the maximum rainfall between the current forecast time and the maximum lead time exceeds the locally defined rainfall warning threshold, a refined flood simulation, and forecast are subsequently performed for this period of time.
\subsection{Hydrodynamic Model}
In a flood event, water depth is generally much smaller than the horizontal inundation extent, and the flow hydrodynamics can be mathematically described by the 2‐D depth‐averaged SWEs~\citep{ming2020real,de2012improving}. SWEs are described by two conservation laws. By neglecting the convective acceleration term, the SWEs for flood modeling  can be written as,
\begin{equation}
\begin{gathered}a
\frac{\partial h}{\partial t}+\frac{\partial q_x}{\partial x}+\frac{\partial q_y}{\partial y}=R-I \\
\frac{\partial q_x}{\partial t}+g h \frac{\partial(h+z)}{\partial x}+\frac{g n^2\left|q\right| q_x}{h^{7 / 3}}=0 \\
\frac{\partial q_y}{\partial t}+g h \frac{\partial(h+z)}{\partial y}+\frac{g n^2\left|q\right| q_y}{h^{7 / 3}}=0,
\end{gathered}
\label{eq44}
\end{equation}
where $t$ is the time index. $x, y$ are the spatial horizontal coordinates. $h$ is the water height relative to the terrain elevation $z$. $q = (q_{x}, q_{y})$ is the discharge per unit width. $R$ represents the rainfall rate, and $I$ is the infiltration rate. It is worth noting that the infiltration rate may not be considered when the infiltration has been basically saturated due to continuous rainfall in the study area.
$g$ is the acceleration due to gravity. $n$ is Manning’s friction coefficient.  
In FloodCast,  the values of Manning’s friction coefficients are based on the range suggested by FLO-2D User’s Manual~\citep{o1993two} and the land cover types of the study area.

To substantially improve the computational efficiency for large‐scale simulations and real‐time forecasting during a flood event induced by intense rainfall, the above governing equations are solved using the proposed sequence-to-sequence GeoPINS.
\subsection{Geometry-adaptive physics-informed neural solver \label{section3}}
\begin{figure}[!htp]
	\centering
	{\includegraphics[width = 1.0\textwidth]{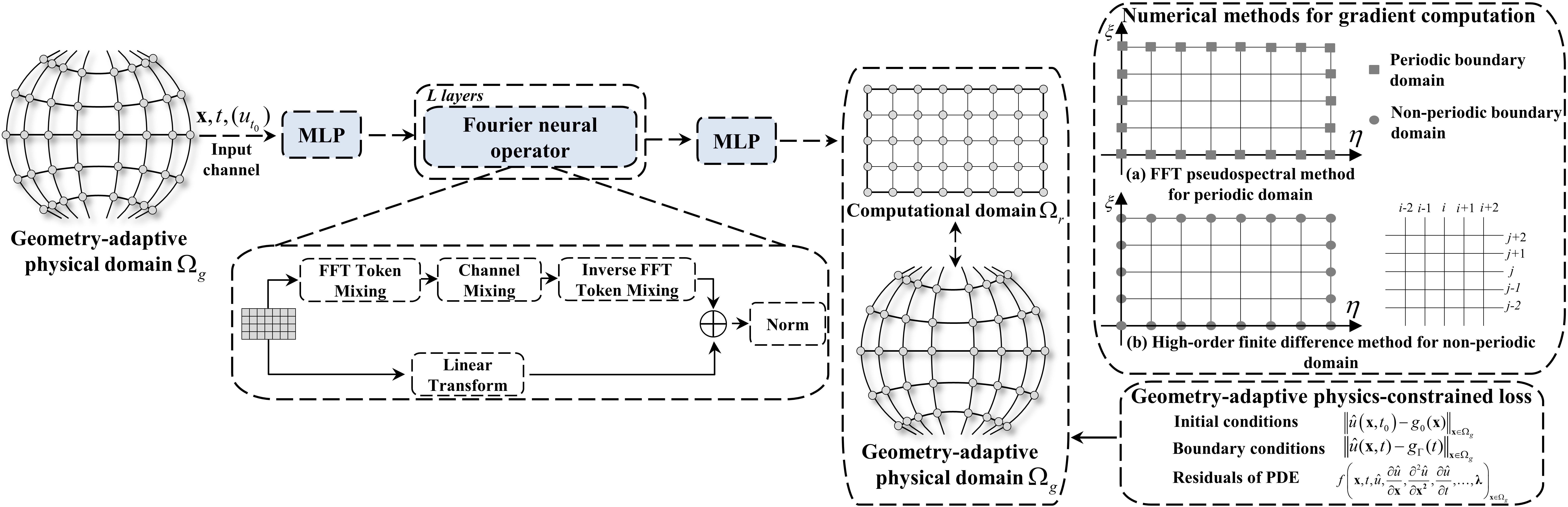}}
	\vspace{-2mm}
	\caption{The full architecture of Geometry-adaptive physics-informed neural solver.}
	\label{fig:2}
	\vspace{-4mm}
\end{figure}

The GeoPINS is built with input channels $I$ composed of coordinates $\mathbf{x}$, time domain $t$ and available initial condition $u_{t_{0}}$ of the physical domain $\Omega_{g}$. As shown in Fig.~\ref{fig:2}, the input is first lifted to a higher dimensional representation $v_{0}$ by multi-layer perceptrons (MLP). Then several FNO layers defined in~\citep{li2020fourier} ($K_{0}$,$K_{1}$,...,$K_{l}$) are applied to extract efficient spatial-temporal representations $v_{l}$.  Where kernel function $\mathcal{K}_{\phi}$:  $\mathbb{R}^{d \times d} \rightarrow \mathbb{R}^{d_{r} \times d_{r}}$ is a neural network parameterized by $\phi$. $d$ represents the dimensionality of discretization.
Notably, different from FNO layers in ~\citep{li2020fourier}, different dimensions of representation space $d_{r}$ in different FNO layers are used to adapt different hydrodynamic problems with lower computational parameters.
The outputs $\hat{u}(t)$ (e.g. height or velocity) in the computational domain $\Omega_{r}$ is obtained by applying MLP projection on $v_{l}$. The forward process can be formulated as,
\begin{equation}
v_{l}=\left(K_{l} \circ \sigma_{l} \circ \cdots \circ \sigma_{1} \circ K_{0}\right) \circ MLP(I), \quad
\hat{u}(t)\|_{\mathbf{x} \in \Omega_{r}}=MLP(v_{l}).
\end{equation}
Here $\hat{u}(\mathbf{x},t)\|_{\mathbf{x} \in \Omega_{r}}$ and $\hat{u}(\mathbf{x},t)\|_{\mathbf{x} \in \Omega_{g}}$ are equal at corresponding nodes.  The neural solver is trained with a proposed Geometry-adaptive physics-constrained (GeoPC) loss constructed by reformulating PINNs in geometry-adaptive space, which is described in detail below.

\subsubsection{Reformulating PINNs in geometry-adaptive space \label{3.1}}
\textbf{Background of PINNs.}
We consider a parametrized PDE system (such as SWEs) given by,
\begin{equation}
\begin{aligned}
&f\left(\mathbf{x}, t, \hat{u}, \frac{\partial \hat{u}}{\partial \mathbf{x}}, \frac{\partial^{2} \hat{u}}{\partial \mathbf{x^{2}}}, \frac{\partial \hat{u}}{\partial t}, \ldots, \vartheta\right)=0, \quad \mathbf{x} \in \Omega, t \in[0, T] \\
&\hat{u}\left(\mathbf{x}, t_{0}\right)=g_{0}(\mathbf{x}) \quad \mathbf{x} \in \Omega \\
&\hat{u}(\mathbf{x}, t)=g_{\Gamma}(t), \quad \mathbf{x} \in \partial \Omega, t \in[0, T]
\end{aligned}
\label{eq1}
\end{equation}
where $\hat{u}: [0, T] \times \mathbb{X} \rightarrow \mathbb{R}^{n}$ is the solution, with initial condition $g_{0}(\mathbf{x})$ and boundary condition $g_{\Gamma}(t)$.
$\vartheta$ = [$\vartheta1$, $\vartheta2$, . . . ] are the PDE parameters. $f$ denotes the residual of the PDE, containing the differential operators (i.e., $ \frac{\partial \hat{u}}{\partial \mathbf{x}}, \frac{\partial^{2} \hat{u}}{\partial \mathbf{x^{2}}}, \frac{\partial \hat{u}}{\partial t}$, . . . ). $\Omega$ and $\partial \Omega$ represent the physical domain and the boundary, respectively.
When the governing PDEs are known, their solutions can be approximated in a physics-informed fashion with less training data~\citep{sun2020physics} or without any training data~\citep{jin2021nsfnets, bihlo2022physics}. 

In PINNs, solving a PDE system (denoted by Eq.~\ref{eq1}) is converted into an optimization problem by minimizing the loss function to iteratively update the neural network (NN),
\begin{equation}
\begin{gathered}
\min _{\theta} \|\mathcal{L}_{\text {data }}\|_{\Omega}+\lambda \|\mathcal{L}_{P D E}\|_{\Omega} \\
\mathcal{L}_{\text {data }}=\|\mathcal{L}_{I C}\|_{\Omega}+\|\mathcal{L}_{B C}\|_{\partial \Omega} \quad
\mathcal{L}_{P D E} = \|f\left(\mathbf{x}, t, \hat{u}, \frac{\partial \hat{u}}{\partial \mathbf{x}}, \frac{\partial^{2} \hat{u}}{\partial \mathbf{x^{2}}}, \frac{\partial \hat{u}}{\partial t}, \ldots, \vartheta\right)\|_{\Omega},
\end{gathered}
\label{eq2}
\end{equation}
where $\lambda$ is a regularization parameter that controls the emphasis on residuals of PDE. $\mathcal{L}_{\text {data }}$ (including $\mathcal{L}_{I C}$/$\mathcal{L}_{B C}$) measure the mismatch between the NN prediction and the initial/boundary conditions in Eq.~\ref{eq1}, and $\mathcal{L}_{P D E}$ is a constraint on the residual of the PDE system. $\theta$ denotes the NN parameters, which takes the spatial and temporal coordinates (x, t), and possibly other quantities, as inputs and then outputs $\hat{u}$. 

\textbf{Reformulating PINNs in geometry-adaptive space.}
Given the geometry-adaptive physical domains $\Omega_{g} = [x,y]$ and computational domains $\Omega_{r} = [\eta,\xi]$,
the transformation $\mathcal{G}$ between coordinates of the geometry-adaptive physical domain ($\Omega_{g}$) and regular computational domain ($\Omega_{r}$) can be defined as,
\begin{equation}
x=\mathcal{G}(\eta,\xi), \quad y=\mathcal{G}(\eta,\xi).
\end{equation}
To enable physics-constrained learning in the geometry-adaptive space, the original PDEs defined in the physical domain (Eq.~\ref{eq1}) have to be recast as the form in computational domain. Specifically,  the first and second derivatives in $\Omega_{g}$ are transformed into $\Omega_{r}$ as,
\begin{equation}
\begin{aligned}
\frac{\partial}{\partial x} &=\left(\frac{\partial}{\partial \eta}\right)\left(\frac{\partial \eta}{\partial x}\right)+\left(\frac{\partial}{\partial \xi}\right)\left(\frac{\partial \xi}{\partial x}\right), \quad
\frac{\partial}{\partial y} =\left(\frac{\partial}{\partial \eta}\right)\left(\frac{\partial \eta}{\partial y}\right)+\left(\frac{\partial}{\partial \xi}\right)\left(\frac{\partial \xi}{\partial y}\right) \\
\frac{\partial^{2}}{\partial x^{2}}&=\frac{\partial \eta}{\partial x} \frac{\partial \eta}{\partial x} \frac{\partial^{2}}{\partial \eta^{2}}+\frac{\partial \eta}{\partial x} \frac{\partial \xi}{\partial x} \frac{\partial^{2}}{\partial \eta \partial \xi}+\frac{\partial \xi}{\partial x} \frac{\partial \xi}{\partial x} \frac{\partial^{2}}{\partial \xi^{2}}+\left[\frac{\partial \xi}{\partial x} \frac{\partial}{\partial \xi}\left(\frac{\partial \eta}{\partial x}\right)+\frac{\partial \eta}{\partial x} \frac{\partial}{\partial \eta}\left(\frac{\partial \eta}{\partial x}\right)\right] \frac{\partial}{\partial \eta} \\
\frac{\partial^{2}}{\partial y^{2}}&=\frac{\partial \eta}{\partial y} \frac{\partial \eta}{\partial y} \frac{\partial^{2}}{\partial \eta^{2}}+\frac{\partial \eta}{\partial y} \frac{\partial \xi}{\partial y} \frac{\partial^{2}}{\partial \eta \partial \xi}+\frac{\partial \xi}{\partial y} \frac{\partial \xi}{\partial y} \frac{\partial^{2}}{\partial \xi^{2}}+\left[\frac{\partial \xi}{\partial y} \frac{\partial}{\partial \xi}\left(\frac{\partial \eta}{\partial y}\right)+\frac{\partial \eta}{\partial y} \frac{\partial}{\partial \eta}\left(\frac{\partial \eta}{\partial y}\right)\right] \frac{\partial}{\partial \eta}.
\end{aligned}
\label{eq4}
\end{equation}
With these modified derivative terms, the GeoPC loss is  defined as,
\begin{equation}
\begin{gathered}
\mathcal{L}_{\text {GeoPC}} = \min _{\theta} \|\mathcal{L}_{\text {data }}\|_{\Omega_{g}}+\lambda \|\mathcal{L}_{P D E}\|_{\Omega_{g}} \\
\mathcal{L}_{\text {data }}=\|\mathcal{L}_{I C}\|_{\Omega_{g}}+\|\mathcal{L}_{B C}\|_{\Omega_{g}} \\
\mathcal{L}_{P D E} = f\left(\mathbf{x}, t, \hat{u}, \frac{\partial \hat{u}}{\partial \mathbf{x}}, \frac{\partial^{2} \hat{u}}{\partial \mathbf{x^{2}}}, \frac{\partial \hat{u}}{\partial t}, \ldots, \vartheta\right), \quad \mathbf{x} \in \Omega_{g}, t \in[0, T].
\end{gathered}
\end{equation}
Two types of gradients in GeoPC loss, including gradients on regular computational domain (e.g. $\frac{\partial}{\partial \eta}$, $\frac{\partial }{\partial \xi}$) and gradients of coordinate transformations (e.g. $\frac{\partial \eta}{\partial x}, \frac{\partial \eta}{\partial y}, \frac{\partial \xi}{\partial x}, \frac{\partial \xi}{\partial y}$), need to be solved efficiently.
The most general way in PINNs~\citep{raissi2019physics,jin2021nsfnets} to compute the exact gradient is to use the auto-differentiation library of neural networks (autograd).
The autograd method is a pointwise operation, which is general and exact. However, it is usually slow and memory-consuming in the space domain. The gradient solving based on numerical methods is significantly more effective than autograd since the number of neural operator parameters is usually much greater than the grid size. Thus, FFT pseudo-spectral (FFT PS) numerical methods~\citep{boyd2001chebyshev} and high-order FD numerical methods~\citep{li1995compact} are utilized to calculate spatial gradients of 
 periodic and non-periodic domains, respectively, which is shown in the upper right corner of Fig.~\ref{fig:2}. The high-order FD methods are used to calculate temporal gradients.

\textbf{Gradients on the computational domain.}
The principal advantage of fast Fourier transformation pseudo-spectral (FFT PS) methods is the great accuracy of the resulting discretization scheme for a given number of nodes, which is the limit of finite-difference methods of increasing orders~\citep{fornberg1987pseudospectral}.  In addition, the saving of computational resources is another advantage of FFT PS methods.  Applying the PS method amounts to performing a periodic discrete convolution, requiring two fast
Fourier transforms.  Specifically, a FFT of the predictions (velocity, depth, and others) $\mathcal{F}(\hat{u})$ in the computational domain is first used to compute the discrete Fourier space.  Then, the solution of gradients can be obtained in the Fourier space. Finally, the inverse FFT $\mathcal{F}^{-1}$ is utilized to construct the full solution in the computational domain. Thus,  
\begin{equation}
\begin{gathered}
\frac{\partial \hat{u}}{\partial \eta}=\mathcal{F}^{-1}\left( ik_{\eta}\mathcal{F}(\hat{u})\right), \quad
\frac{\partial^{2} \hat{u}}{\partial \eta^{2}}=\mathcal{F}^{-1}\left( -k_{\eta}^{2}\mathcal{F}(\hat{u})\right), \quad
k_{\eta}=\frac{2 \pi w_{\eta}}{L_{\eta}} \\ 
\frac{\partial \hat{u}}{\partial \xi}=\mathcal{F}^{-1}\left( ik_{\xi}\mathcal{F}(\hat{u})\right), \quad  \frac{\partial^{2} \hat{u}}{\partial \xi^{2}}=\mathcal{F}^{-1}\left( -k_{\xi}^{2}\mathcal{F}(\hat{u})\right), \quad
k_{\xi}=\frac{2 \pi w_{\xi}}{L_{\xi}} \\
w_{\eta} = \{-N_{\eta} / 2,-N_{\eta} / 2+1, \ldots, N_{\eta} / 2-1\} \quad \\ w_{\xi}=\{-N_{\xi} / 2,-N_{\xi} / 2+1, \ldots, N_{\xi} / 2-1\}, 
\end{gathered}
\end{equation}
where $k_{\eta}$ and $k_{\xi}$ represent  $\eta$-direction wave-numbers and $\xi$-direction wave-numbers, respectively.  The spatial grid size of the computational domain is $N_{\eta} \times N_{\xi}$.  $L_{\eta}$ and $L_{\xi}$ are spatial lengths of $\eta$-direction and $\xi$-direction, respectively. Importantly, if rapidly-oscillating PDE functions with relatively few grid points are solved (e.g. Convection equation with spatial size $32\times32$), an effect known as aliasing will be observed~\citep{boyd2001chebyshev}.  Aliasing means that wavenumbers of magnitude greater than $k_{\eta}/2$ or $k_{\xi}/2$ are incorrectly represented as lower wavenumbers. A 2/3 filter~\citep{hou2007computing} is used to avoid aliasing instability. 
Filtering means removing the mode that leads to aliasing, which can be done by damping the high wavenumbers when computing the gradients. 

However, FFT PS methods are restricted to computation in rectangular domains with periodic boundary conditions. On the contrary, some PS solutions are designed for non-periodic boundary conditions. For example,  the non-periodic boundary (e.g. no-slip wall condition) can be handled by Chebyshev PS methods~\citep{farcy1988chebyshev}. However, a non-uniform collocation grid (e.g. cosine function) needs to be built for Chebyshev polynomials, which makes  FD-based gradient computation of coordinate transformations unstable and introduces construction problems of the differential scheme. Therefore, a four-order difference scheme is constructed to obtain stable and accurate gradients on the computational domain with a non-periodic boundary. Specifically, the first derivatives are approximated by four-order central differences,
\begin{equation}
\begin{aligned}
\frac{\partial \hat{u}}{\partial \eta} & \approx \frac{-\hat{u}_{\eta+2, \xi}+8 \hat{u}_{\eta+1, \xi}-8 \hat{u}_{\eta-1, \xi}+\hat{u}_{\eta-2, \xi}}{12 \delta_{\eta}}+O\left((\delta_{\eta})^{4}\right) \\
\frac{\partial \hat{u}}{\partial \xi} & \approx \frac{-\hat{u}_{\eta, \xi+2}+8 \hat{u}_{\eta, \xi+1}-8 \hat{u}_{\eta, \xi-1}+\hat{u}_{\eta, \xi-2}}{12 \delta_{\xi}}+O\left((\delta_{\xi})^{4}\right),
\end{aligned}
\label{eq9}
\end{equation}
where $\delta_{\eta}$ and $\delta_{\xi}$ are spatial steps of  $\eta$-direction and $\xi$-direction, respectively. The four-order difference can be expressed by a convolution filter in the algorithm. 
For the boundary, third-order one-sided differences (i.e., upwind/downwind) are applied to stabilize the difference scheme.

 \textbf{Gradients of coordinate transformations.}
The gradient solutions of coordinate transformations  $\mathcal{G}: \Omega_{r} \mapsto \Omega_{g}$  
can be divided into two cases. Firstly,  when analytical forms of $\mathcal{G}$ are available, gradients of coordinate transformations can be solved exactly. For example, in terms of solving the Convection equation and  SWEs for flood modeling in regular rectangular grid ($\Omega_{r} = \Omega_{g}$), gradients of coordinate transformations are expressed as,
\begin{equation}
\left[\begin{array}{cc}
\frac{\partial \eta}{\partial x} & \frac{\partial \eta}{\partial y} \\
\frac{\partial \xi}{\partial x} & \frac{\partial \xi}{\partial y} 
\end{array}\right]=\left[\begin{array}{cc}
1 & 0 \\
0 & 1
\end{array}\right]
\label{eq10}
\end{equation}

Secondly, in some cases, analytical forms of $\mathcal{G}$ are not available, which have to be approximated numerically (e.g. solving incompressible Navier-Stokes equations on an irregular domain). Inspired by~\citep{gao2021phygeonet}, elliptic coordinate transformation is utilized in this case. For example, $\mathcal{G}$ can be obtained by solving a diffusion equation, considering the one-to-one mapping boundary condition. Formulated as,
\begin{equation}
\begin{aligned}
& \alpha \frac{\partial^{2} x}{\partial \eta^{2}}-2 \beta \frac{\partial^{2} x}{\partial \eta \partial \xi}+\gamma \frac{\partial^{2} x}{\partial \xi^{2}}=0, \quad
\alpha \frac{\partial^{2} y}{\partial \eta^{2}}-2 \beta \frac{\partial^{2} y}{\partial \eta \partial \xi}+\gamma \frac{\partial^{2} y}{\partial \xi^{2}}=0 \\
& \alpha =\left(\frac{\partial x}{\partial \xi}\right)^{2}+\left(\frac{\partial y}{\partial \xi}\right)^{2}, 
\gamma =\left(\frac{\partial x}{\partial \eta}\right)^{2}+\left(\frac{\partial y}{\partial \eta}\right)^{2},
\beta =\frac{\partial x}{\partial \eta} \frac{\partial x}{\partial \xi}+\frac{\partial y}{\partial \eta} \frac{\partial y}{\partial \xi} \\
& Boundary: \mathcal{G}(\eta,\xi)=\partial \Omega_{g}^{i} \quad for \quad  \forall \eta,\xi \in \partial \Omega_{r}^{i}, \quad i=1, \cdots, 4.
\end{aligned}
\label{eq.11}
\end{equation}
By solving Eq.~\ref{eq.11} based on four-order difference numerically (Eq.~\ref{eq9}), the Jacobian  $\frac{\partial y}{\partial \eta}, \frac{\partial y}{\partial \xi}, \frac{\partial x}{\partial \eta}$ and $\frac{\partial x}{\partial \xi}$ of $\mathcal{G}$ can be obtained. Then, the  Jacobian matrix  ($J=\frac{\partial x}{\partial \eta} \frac{\partial y}{\partial \xi}-\frac{\partial x}{\partial \xi} \frac{\partial y}{\partial \eta}$) is applied to modify Eq.~\ref{eq4} (such as first derivatives) as,
\begin{equation}
\begin{aligned}
\frac{\partial}{\partial x}&=J^{-1}\left[\left(\frac{\partial}{\partial \eta}\right)\left(\frac{\partial y}{\partial \xi}\right)-\left(\frac{\partial}{\partial \xi}\right)\left(\frac{\partial y}{\partial \eta}\right)\right] \\
\frac{\partial}{\partial y}&=J^{-1}\left[\left(\frac{\partial}{\partial \xi}\right)\left(\frac{\partial x}{\partial \eta}\right)-\left(\frac{\partial}{\partial \eta}\right)\left(\frac{\partial x}{\partial \xi}\right)\right].
\end{aligned}
\label{eq13}
\end{equation}

The derivatives of reference domains $\Omega_{r} = [\eta,\xi]$ with respect to geometry-adaptive physical domains $\Omega_{g} = [x,y]$ can be expressed as,
\begin{equation}
\begin{aligned}
\frac{\partial \eta}{\partial x} &=\frac{1}{J} \frac{\partial y}{\partial \xi}, \quad
\frac{\partial \xi}{\partial x} &=-\frac{1}{J} \frac{\partial y}{\partial \eta}, \quad
\frac{\partial \eta}{\partial y} &=-\frac{1}{J} \frac{\partial x}{\partial \xi}, \quad
\frac{\partial \xi}{\partial y} &=\frac{1}{J} \frac{\partial x}{\partial \eta},
\end{aligned}
\label{eq11}
\end{equation}
where  the  Jacobian matrix  $J=\frac{\partial x}{\partial \eta} \frac{\partial y}{\partial \xi}-\frac{\partial x}{\partial \xi} \frac{\partial y}{\partial \eta}$. 

The second derivatives in $\Omega_{g}$ are transformed into $\Omega_{r}$ as,
\begin{equation}
\begin{aligned}
\frac{\partial^{2}}{\partial x^{2}}&=\frac{\partial \eta}{\partial x} \frac{\partial \eta}{\partial x} \frac{\partial^{2}}{\partial \eta^{2}}+\frac{\partial \eta}{\partial x} \frac{\partial \xi}{\partial x} \frac{\partial^{2}}{\partial \eta \partial \xi}+\frac{\partial \xi}{\partial x} \frac{\partial \xi}{\partial x} \frac{\partial^{2}}{\partial \xi^{2}}+\left[\frac{\partial \xi}{\partial x} \frac{\partial}{\partial \xi}\left(\frac{\partial \eta}{\partial x}\right)+\frac{\partial \eta}{\partial x} \frac{\partial}{\partial \eta}\left(\frac{\partial \eta}{\partial x}\right)\right] \frac{\partial}{\partial \eta} \\
\frac{\partial^{2}}{\partial y^{2}}&=\frac{\partial \eta}{\partial y} \frac{\partial \eta}{\partial y} \frac{\partial^{2}}{\partial \eta^{2}}+\frac{\partial \eta}{\partial y} \frac{\partial \xi}{\partial y} \frac{\partial^{2}}{\partial \eta \partial \xi}+\frac{\partial \xi}{\partial y} \frac{\partial \xi}{\partial y} \frac{\partial^{2}}{\partial \xi^{2}}+\left[\frac{\partial \xi}{\partial y} \frac{\partial}{\partial \xi}\left(\frac{\partial \eta}{\partial y}\right)+\frac{\partial \eta}{\partial y} \frac{\partial}{\partial \eta}\left(\frac{\partial \eta}{\partial y}\right)\right] \frac{\partial}{\partial \eta}.
\end{aligned}
\label{eq22}
\end{equation}
Based on Eq.~\ref{eq11} and Eq.~\ref{eq22}, the second derivatives are calculated as,
\begin{equation}
\begin{aligned}
\frac{\partial^{2}}{\partial x^{2}}&=J^{-2}\left[\frac{\partial y}{\partial \xi} \frac{\partial y}{\partial \xi} \frac{\partial^{2}}{\partial \eta^{2}}-\frac{\partial y}{\partial \xi} \frac{\partial y}{\partial \eta} \frac{\partial^{2}}{\partial \eta \partial \xi}+\frac{\partial y}{\partial \eta} \frac{\partial y}{\partial \eta} \frac{\partial^{2}}{\partial \xi^{2}}+\left[-\frac{\partial y}{\partial \eta} \frac{\partial}{\partial \xi}\left(\frac{\partial y}{\partial \xi}\right)+\frac{\partial y}{\partial \xi} \frac{\partial}{\partial \eta}\left(\frac{\partial y}{\partial \xi}\right)\right] \frac{\partial}{\partial \eta}\right] \\
\frac{\partial^{2}}{\partial y^{2}}&=J^{-2}\left[\frac{\partial x}{\partial \xi} \frac{\partial x}{\partial \xi} \frac{\partial^{2}}{\partial \eta^{2}}-\frac{\partial x}{\partial \xi} \frac{\partial x}{\partial \eta} \frac{\partial^{2}}{\partial \eta \partial \xi}+\frac{\partial x}{\partial \eta} \frac{\partial x}{\partial \eta} \frac{\partial^{2}}{\partial \xi^{2}}+\left[-\frac{\partial x}{\partial \eta} \frac{\partial}{\partial \xi}\left(\frac{\partial x}{\partial \xi}\right)+\frac{\partial x}{\partial \xi} \frac{\partial}{\partial \eta}\left(\frac{\partial x}{\partial \xi}\right)\right] \frac{\partial}{\partial \eta}\right].
\end{aligned}
\label{eq33}
\end{equation}
\subsubsection{Spatial-temporal model \label{3.2}}
FNO is an effective spatial-temporal model to resolve the issues of PINNs.  The discrete FNO is defined by replacing the kernel function in~\citep{li2020fourier} with, 
\begin{equation}
\left(\mathcal{K}(\phi) v_{t}\right)(x) = \mathcal{F}^{-1}\left(R_{\phi} \cdot\left(\mathcal{F} v_{t}\right)\right)(x) \quad \forall x \in \Omega,
\label{eq14}
\end{equation}
where $\mathcal{F}$ and $\mathcal{F}^{-1}$ represent the FFT and inverse FFT, respectively. $R_{\phi}$  is  a parametric function $\mathbb{R}^{d} \times \mathbb{R}^{d_{r}} \rightarrow \mathbb{R}^{d_r} \times \mathbb{R}^{d_{r}}$ that maps to the values of the appropriate Fourier modes $k$.
Two perspectives are given to theoretically analyze the effectiveness of FNO. 

\textbf{Fourier feature embeddings.}
FNO may enable fast convergence to high-frequency components of PDEs.  Conventional fully connected neural networks (FCNN) in PINNs can yield a significantly lower convergence rate for high-frequency components of PDEs~\citep{wang2021eigenvector}. However, compared with FCNN, Fourier feature embeddings by FFT in Eq.~\ref{eq14} can make networks better suited for modeling PDEs' systems in low dimensions, thereby overcoming the high-frequency failure problems of PINNs, which can be demonstrated by neural tangent kernel theory~\citep{tancik2020fourier}. 
 
\textbf{FNO is an effective vision mixer.}
Vision transformers (ViT) have recently shown promise in producing rich contextual representations for spatial-temporal tasks~\citep{dosovitskiy2020image, yu2021metaformer}. 
The general architecture of ViT~\citep{dosovitskiy2020image} is a line of models with a stack of vision mixers,  which include token mixing $\rightarrow$ channel mixing $\rightarrow$ token mixing.
Mixer components can be various attention-based operations~\citep{dosovitskiy2020image}, spatial MLP~\citep{tolstikhin2021mlp}, pooling  operation~\citep{yu2021metaformer}, and others.
A key component for the effectiveness of transformers is attributed to the proper mixing of tokens. 

The FNO in Eq.~\ref{eq14} is a member of vision mixers. Specifically, the FNO can be decomposed into token mixing $\rightarrow$ channel mixing $\rightarrow$ token mixing. Firstly, the FFT $\mathcal{F} v_{t}$ can be considered the token mixing. Then, the matrix multiplication $R_{\phi}$ performs mixing on the channel. Finally, the inverse FFT $\mathcal{F}^{-1}$ is a token mixing layer. Detailed process is shown in Fig.~\ref{fig:2}. Thus, FNO has superior spatiotemporal expressiveness and generalization for challenging PDEs.

\subsubsection{Numerical experiments  \label{section4}}
In this section, we conduct empirical experiments to examine the efficacy of the proposed GeoPINS by solving the 1-D Convection equation on regular domains, and 2-D incompressible Navier-Stokes equation on irregular river channels.

\textbf{Implementation details.} Four FNO layers are stacked with different Fourier modes $k$ = \{12, 12, 9, 9\}, channel combinations of  the representation space $d_{r}$ = \{16, 24, 24, 32, 32\} and GeLU activation for all cases. A layer normalization is used after the FNO layer. We use Adam optimizer~\citep{kingma2014adam}  with an initial learning rate of 0.001. The hyperparameter $\lambda$=1 in GeoPC loss for numerical experiments.
$L2$ relative error between the predicted solutions and the analytical/computational fluid dynamics (CFD) solutions is regarded as an evaluation measure. For all cases, we run models at least five times with different preset random seeds and average the relative errors. 

\textbf{Benchmarks.} State-of-the-art solvers based on PINNs are considered as benchmarks, including 
regular PINN (a 4-layer fully-connected NN~\citep{krishnapriyan2021characterizing}), Physics-informed geometry-adaptive convolutional neural networks (PhyGeoNet)~\citep{gao2021phygeonet}, 
PINO~\citep{li2021physics}, and PINN-DeepONet~\citep{wang2021learning}.

\begin{table}[!htp]
	\caption{Results of Convection equations. $A$ $\rightarrow$$B$ represents training on a spatial resolution with $A$ and directly evaluating on a spatial resolution with $B$. The values in \textbf{bold} are the best.}
	\centering
	\resizebox{1.05\textwidth}{!}
	{
		\centering
		\begin{tabular}{c|c|cccc|cccc|cccc}
			\hline
			\multirow{2}{*}{Methods} &\multirow{2}{*}{Mode Parameters} & \multicolumn{4}{c|}{$\beta$=20}                                               & \multicolumn{4}{c|}{$\beta$=30}                                               & \multicolumn{4}{c}{$\beta$=40}                                                \\ \cline{3-14} 
			& & 32$\rightarrow$256          & 64$\rightarrow$256          & 128$\rightarrow$256         & 256$\rightarrow$256         & 32$\rightarrow$256          & 64$\rightarrow$256          & 128$\rightarrow$256         & 256$\rightarrow$256         & 32$\rightarrow$256          & 64$\rightarrow$256          & 128$\rightarrow$256         & 256$\rightarrow$256         \\ \hline
			Regular PINN (a 4-layer fully-connected NN) &  30701          & 1.6667          & 1.7081          & 1.5183          & 1.1276          & 1.7325          & 1.9008          & 1.5630          & 1.0372          & 1.5085          & 1.8210          & 1.0184          & 1.0993          \\ \hline
			Regular PINN (a 8-layer fully-connected NN) &  1124401          & 1.2277          & 1.4874          & 0.9537          & 1.3654          & 0.9569          & 1.4404          & 1.3859          & 1.0723          & 1.4620          & 1.1529          & 1.1012          & 1.2013          \\ \hline
			GeoPINS with FD   &   1141091       & 0.3128          & 0.0974          & \textbf{0.0271} & 0.0770          & 0.4323          & 0.1308          & \textbf{0.0332} & 0.1077          & 0.5553          & \textbf{0.1660} & \textbf{0.0524} & \textbf{0.2310} \\ \hline
			GeoPINS with FFT   PS  & 1141091    & \textbf{0.0622} & \textbf{0.0381} & 0.0304          & \textbf{0.0225} & \textbf{0.1608} & \textbf{0.1211} & 0.1035          & \textbf{0.0952} & \textbf{0.4668} & 0.4436          & 0.4296          & 0.4215          \\ \hline
		\end{tabular}
	}
	\label{Tab:Table1}
\end{table}
\textbf{Convection equation on regular domain.} We consider a 1-D convection problem to verify the proposed GeoPINS, which is a common hyperbolic wave equation. It takes the form,
\begin{equation}
\begin{aligned}
\frac{\partial u}{\partial t}+\beta \frac{\partial u}{\partial x} &=0, \quad x \in[0, 2 \pi], t \in[0, T] \\
u(x, 0) &=h(x),
\end{aligned}
\label{eq.16}
\end{equation}
with periodic boundary conditions $u(0, t)=u(2 \pi, t)$. Where $\beta$ is the convection coefficient and $h(x)$ is the initial condition. $h(x)=\sin (x)$ in our experiments. An analytical solution of this problem is used to verify the prediction of neural-based solvers,
\begin{equation}
u_{\text {analytical }}(x, t)=\mathcal{F}^{-1}\left(\mathcal{F}(h(x)) e^{-i \beta k t}\right),
\end{equation}
where $k$ denotes frequency in the Fourier domain. The GeoPC loss is defined by combining Eq.~\ref{eq.16} and Eq.~\ref{eq10},
\begin{equation}
\begin{gathered}
\mathcal{L}_{\text {GeoPC}} = \min _{\theta} \|\hat{u}(x, 0)-u(x, 0)\| + \|\hat{u}(0, t)-\hat{u}(2 \pi, t)\| + \lambda\|\frac{\partial \hat{u}}{\partial t}+\beta \frac{\partial \hat{u}}{\partial x}^{2}\|, \\
x \in[0, 2 \pi], t \in[0, T].
\end{gathered}
\end{equation}
\begin{figure}[!tp]
	\centering
	{\includegraphics[width = 1.05\textwidth]{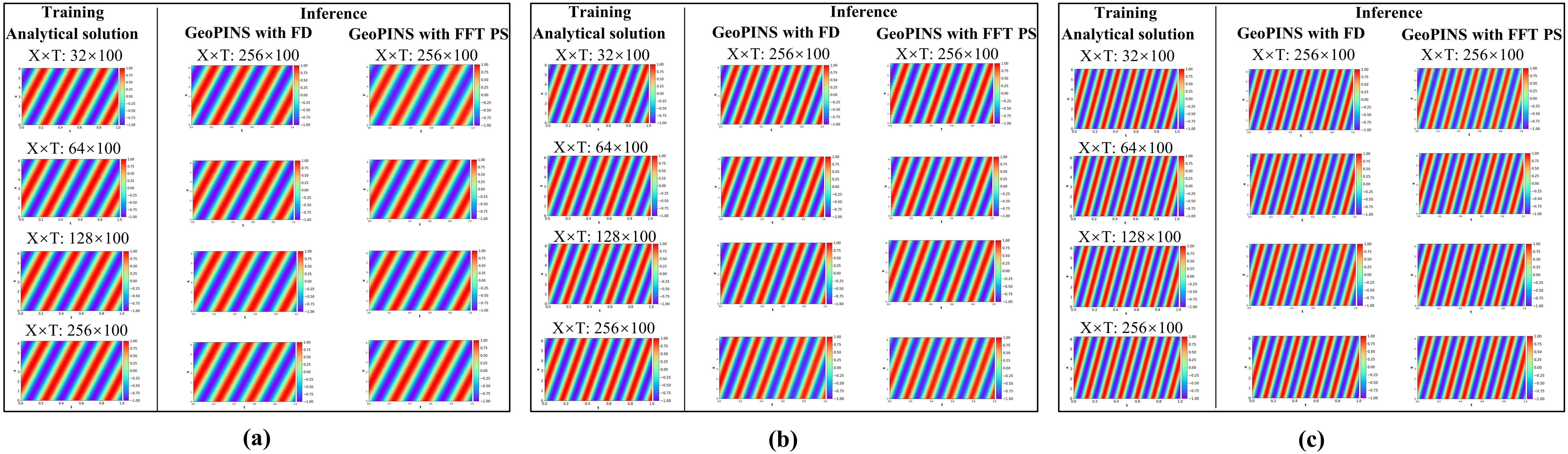}}
        \vspace{-2mm}
	\caption{Train on a  low spatial-temporal resolution (first column) directly infer on a  spatial-temporal resolution $256 \times 100$ by GeoPINS with FD (second column) and GeoPINS with FFT PS (third column). (a) 1-D convection equations on regular domain ($\beta$=20). (b) 1-D convection equations on regular domain ($\beta$=30). (c) 1-D convection equations on regular domain ($\beta$=40)}
	\label{fig:11}
\end{figure}
We use batch size = 1. Adam optimizer is utilized with a learning rate of 0.001 that decays by half every 50 epochs. 20000 epochs in total. Each epoch has only one iteration step. The input channel is set to  3, including coordinate $x$, time $t$, and initial condition $u_{0}$. 

The results of our experiments are shown in Table~\ref{Tab:Table1}. As shown in Table~\ref{Tab:Table1}, GeoPINS with FD or FFT PS results in a much more accurate solution than the regular PINN. With GeoPINS, the relative error is almost two orders of magnitude lower under different zero-shot super-resolution settings. Furthermore, the error is almost invariant for GeoPINS with different resolutions. 
Compared with  GeoPINS with FD, GeoPINS with FFT PS is generally better performing for this problem with periodic boundary, especially for extreme zero-shot super-resolution (training on spatial-temporary resolution $32 \times 100$ and directly inferring on $256 \times 100$). However, GeoPINS with FD is more robust for solving complex and non-symmetric convection equations (e.g. $\beta$=40).  Visualized results of different convection coefficients ($\beta$=20,30,40) are shown in Fig.~\ref{fig:11}. Based on these visualize results, it has been demonstrated that GeoPINS can do zero-shot super-resolution in the spatial domain, and the results are accurate (similar to analytical solutions) under different convection coefficients.

\begin{figure}[!htp]
	\centering
	{\includegraphics[width = .85\textwidth]{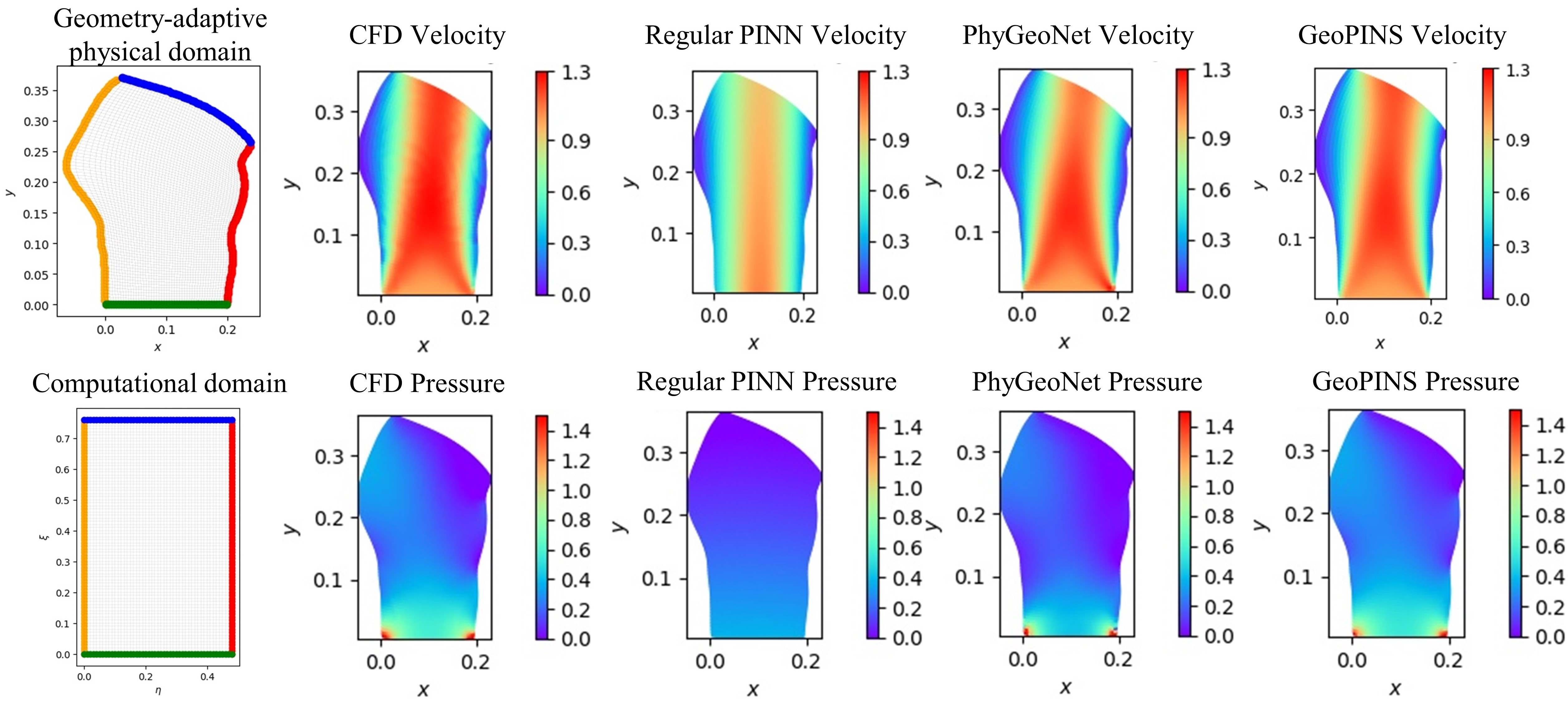}}
	\vspace{-2mm}
	\caption{The physical domain (upper left) and reference domain (lower left) for 2-D Navier-Stokes equations. Velocity (the first row) and pressure (the second row) contours of CFD benchmakrk, Regular PINN, PhyGeoNet, and GeoPINS. All methods are fully trained at $\nu=0.01$.}
	\label{fig:3}
	\vspace{-4mm}
\end{figure}
\begin{table}[!htp]
	\caption{Results of 2-D incompressible Navier-Stokes equations. The values in \textbf{bold} are the best. It is demonstrated that GeoPINS has excellent geometry-adaptive and resolution-invariant properties.}
	\centering
	\resizebox{0.98\textwidth}{!}
	{   \begin{tabular}{c|ccc|ccc}
			\hline
			\multirow{2}{*}{Methods} & \multicolumn{3}{c|}{25$\times$39 $\rightarrow$ 49$\times$77}                & \multicolumn{3}{c}{49$\times$77$\rightarrow$ 49$\times$77}                 \\ \cline{2-7} 
			& Training time per epoch & Velocity      & Pressure     & Training time per epoch & Velocity      & Pressure      \\ \hline
			Regular PINN                    & 0.0313                  & 0.2864 & 0.6866 & 0.1571                  & 0.2375 & 0.6326 \\
			PhyGeoNet                & 0.0615                  & 0.2391 & 0.6551 & 0.0640                  & 0.0858 & 0.3400 \\ \hline
			GeoPINS                  & 0.0648                  & \textbf{0.0993} &\textbf{ 0.3681} & 0.0669                  & \textbf{0.0611} & \textbf{0.2769} \\ \hline
		\end{tabular}
	}
	\label{Tab:Table3}
\end{table}
\textbf{Navier-Stokes equation on irregular river channel.}
We consider the 2-d steady incompressible Navier-Stokes equations,
\begin{equation}
\nabla \cdot \mathbf{v} =0, \quad
(\mathbf{v} \cdot \nabla) \mathbf{v}=-\nabla p + \nabla \cdot(\nu \nabla \mathbf{v})
\label{eq20}
\end{equation}
where $\mathbf{v}$, $p$ represent the velocity vector and pressure, respectively. $\nu$ is fluid viscosity. The fluid flow is solved on a geometry-adaptive physical domain $\Omega_{g}$, as shown in the top left of Fig.~\ref{fig:3}. Similar to ~\citep{gao2021phygeonet},  the no-slip wall boundary condition is used on the left and right boundaries, and the boundary condition of the inlet (green)  is given by $v = [0, 1]$. The boundary condition of the outlet (blue)  is set as a symmetrical boundary and $p = 0$. The computational domain $\Omega_{r}$ is a rectangular grid with step =  0.01 (bottom left in Fig.~\ref{fig:3}).

The GeoPC loss can be obtained by combing Eq.~\ref{eq20} and Eq.~\ref{eq13},
\begin{equation}
\begin{gathered}
\mathcal{L}_{\text {GeoPC}} = \min _{\theta} \|\mathcal{L}_{\text {boundary }}\|_{\Omega_{g}}+\lambda \|\nabla \cdot \hat{\mathbf{v}} + (\hat{\mathbf{v}} \cdot \nabla) \hat{\mathbf{v}} + \nabla \hat{p} - \nabla \cdot(\nu \nabla \hat{\mathbf{v}})\|_{\Omega_{g}} 
\end{gathered}
\end{equation}
In terms of implementation details of the Navier-Stokes equation, batch size = 1,  total epochs = 20000. We use Adam optimizer with a learning rate of 0.001 that decays by half every epoch. To evaluate the learning performance,  the ground truth (same with PhyGeoNet~\citep{gao2021phygeonet}) is generated by FV-based numerical simulations. The FV simulations are performed on an open-source FV platform, OpenFOAM~\citep{jasak2007openfoam}. The input channel is set to  2, including the $x$-component coordinate and $y$-component coordinate. 

The visual results of our experiments are shown in Fig.~\ref{fig:3}, where the CFD results are used as a benchmark. The regular PINN fails to accurately predict the velocity and pressure distributions, which shows the flaws of PINNs in multiphysics problems. Both the PhyGeoNet and GeoPINS with FD can capture the flow pattern. However, discrepancies are observed in the velocity outlet region and in the high-pressure region in the left and right corners. The GeoPINS prediction result is more accurate and agrees with the CFD benchmark very well. The relative errors for velocity and pressure are listed in Table~\ref{Tab:Table3}, GeoPINS achieves robust and best performances under different zero-shot super-resolution settings.  Contrary to the failure of PhyGeoNet in the setting 25$\times$39 $\rightarrow$ 49$\times$77, GeoPINS retains the resolution-invariant properties under similar training time as PhyGeoNet. In addition, training time is shorter at lower resolutions. Thus, GeoPINS provides a favorable opportunity for fast and refined-resolution solutions of large-scale PDEs through the geometry-adaptive and resolution-invariant properties.


\subsection{Sequence-to-sequence GeoPINS for Large-scale Flood Modeling}

\begin{figure}[!htp]
	\centering
	{\includegraphics[width = .95\textwidth]{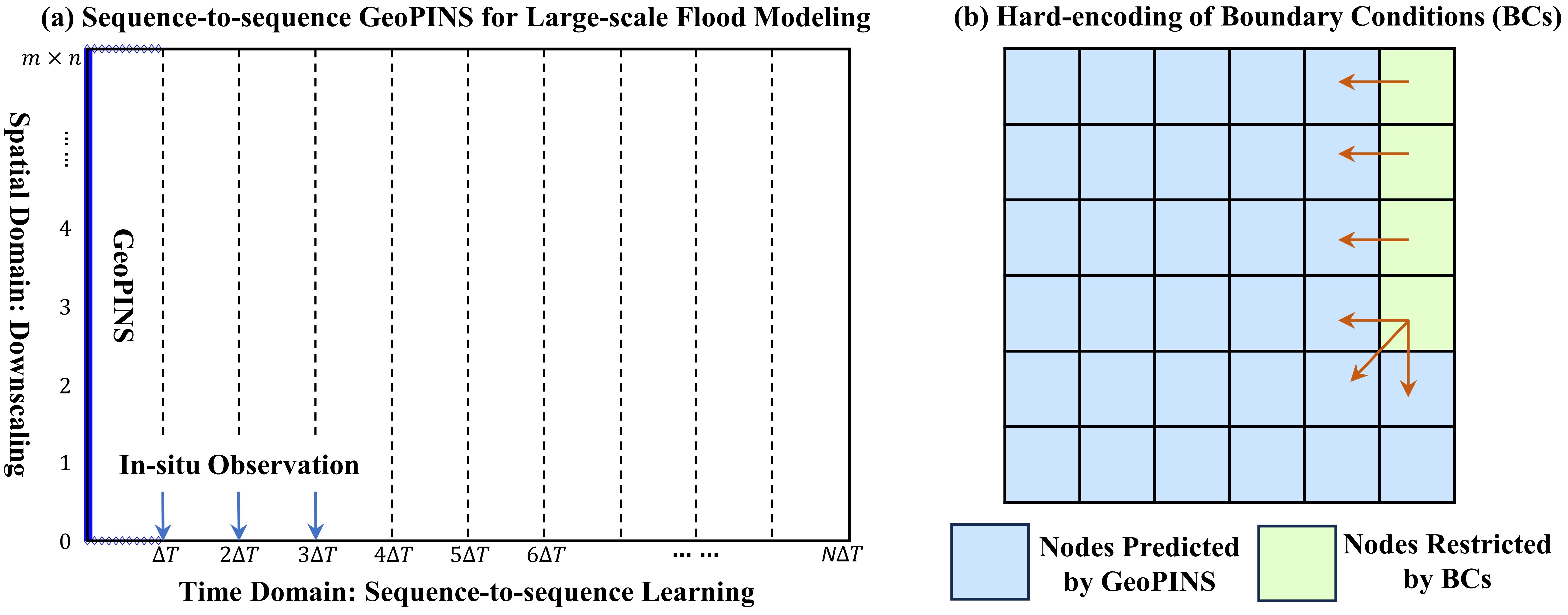}}
	\vspace{-2mm}
	\caption{(a) Schematic architecture of sequence-to-sequence GeoPINS for large-scale flood modeling; (b) Embedding of BCs into the predicted output of GeoPINS.}
	\label{fig:6}
	\vspace{-4mm}
\end{figure}

Large-scale flood modeling and forecasting are inherently distinguished by the presence of long-term temporal series and extensive spatial domains. In order to address these two challenges, a sequence-to-sequence GeoPINS model is proposed for large-scale flood modeling, as shown in Fig.~\ref{fig:6}. Specifically, for tackling large-scale temporal or spatial problems, based on the resolution-invariant feature of GeoPINS, we train GeoPINS on a lower spatio-temporal resolution and directly infer at higher resolution for flood forecast. The original GeoPINS approach trains the NN model to predict the entire space-time, which imposes limitations on modeling long time series dimensions. Therefore, inspired by the sequence-to-sequence learning task, we model long-term sequences into different sequences. A marching-in-time scheme (shown in Fig.~\ref{fig:6} (a)) is utilized to predict different sequences. We take the prediction at $T = \Delta T$ and use this as the initial condition to make a prediction at $T = 2 \Delta T$. In the sequence-to-sequence GeoPINS approach, the initial conditions for the first sequence $\Delta T$ are derived from Earth observation data (such as SAR-based water mapping). However, a significant challenge in this methodology is the potential accumulation of errors in initial conditions. This means that errors in the prediction results of the final time step in one time series may accumulate and affect the initial conditions for the subsequent sequence. To address this issue, we employ real-time or in-situ observation data. Specifically, in the several sequences of the model, actual observations or in-situ measurements are utilized as the real initial conditions for sequence-to-sequence GeoPINS. In-situ observations, such as data recorded by hydrological stations, typically have a small time interval. If we are only able to access in-situ observations at larger time intervals, such as daily data, we can adapt the model by adjusting the time length and time step of each sequence. These in-situ observations enhance the accuracy and reliability of the model's predictions by reducing the errors of initial condition errors.

In order to effectively train the sequence-to-sequence GeoPINS model, the boundary conditions (BCs) (e.g., river inflow boundary) can be encoded into the learning model. Inspired by the idea from the FD method, we apply a hard-encoding of BCs to the model’s prediction at each time step, as shown by Fig.~\ref{fig:6} (b). Specifically, for the river inflow boundary, we change the prediction results at the boundary with prescribed values.  The physical domain $\Omega_{g}$ and computational domain $\Omega_{r}$ is a rectangular grid.

The GeoPC loss can be obtained based on 2-D depth-averaged SWEs (Eq.~\ref{eq44}),
\begin{equation}
\begin{gathered}
\mathcal{L}_{\text {GeoPC}} = \lambda \|\mathcal{L}_{I C}\|_{\Omega_{g}} + \|\mathcal{L}_{P D E}\|_{\Omega_{g}} \\
\mathcal{L}_{P D E} = \|\frac{\partial h}{\partial t}+\frac{\partial q_x}{\partial x}+\frac{\partial q_y}{\partial y}-R+I\| \\
+  \| \frac{\partial q_x}{\partial t}+g h \frac{\partial(h+z)}{\partial x}+\frac{g n^2\left|q\right| q_x}{h^{7 / 3}} \| 
\\ +
\| \frac{\partial q_y}{\partial t}+g h \frac{\partial(h+z)}{\partial y}+\frac{g n^2\left|q\right| q_y}{h^{7 / 3}} \|, \quad x, y \in \Omega_{g}, t \in[0, \Delta T].
\end{gathered}
\end{equation}
Where $\lambda$ is a hyperparameter for balancing the loss, and $\lambda$ is set to 10 by default in our experiments. FD is employed to implement the time or spatial gradients in the equation, primarily due to the presence of aperiodic boundary conditions.
For each sequence of sequence-to-sequence GeoPINS, batch size = 1,  epochs = 1000, $\Delta T$=16. To capture the requisite spatiotemporal patterns, four FNO layers are arranged in a stack, employing spatial Fourier modes represented as $k_{spa} = \{32, 32, 32, 32\}$ and temporal Fourier modes represented as $k_{tem} = \{4, 4, 4, 4\}$. Adam optimizer is utilized with a learning rate of 0.001 that decays by half every 500 epochs. The input channel is set to  3, including coordinate $x$, time $t$, and initial condition. 
 
\section{Study Area and Data}
\subsection{Description of the Study Area}
\begin{figure}[!htp]
\vspace{-2mm}
	\centering
	{\includegraphics[width = .68\textwidth]{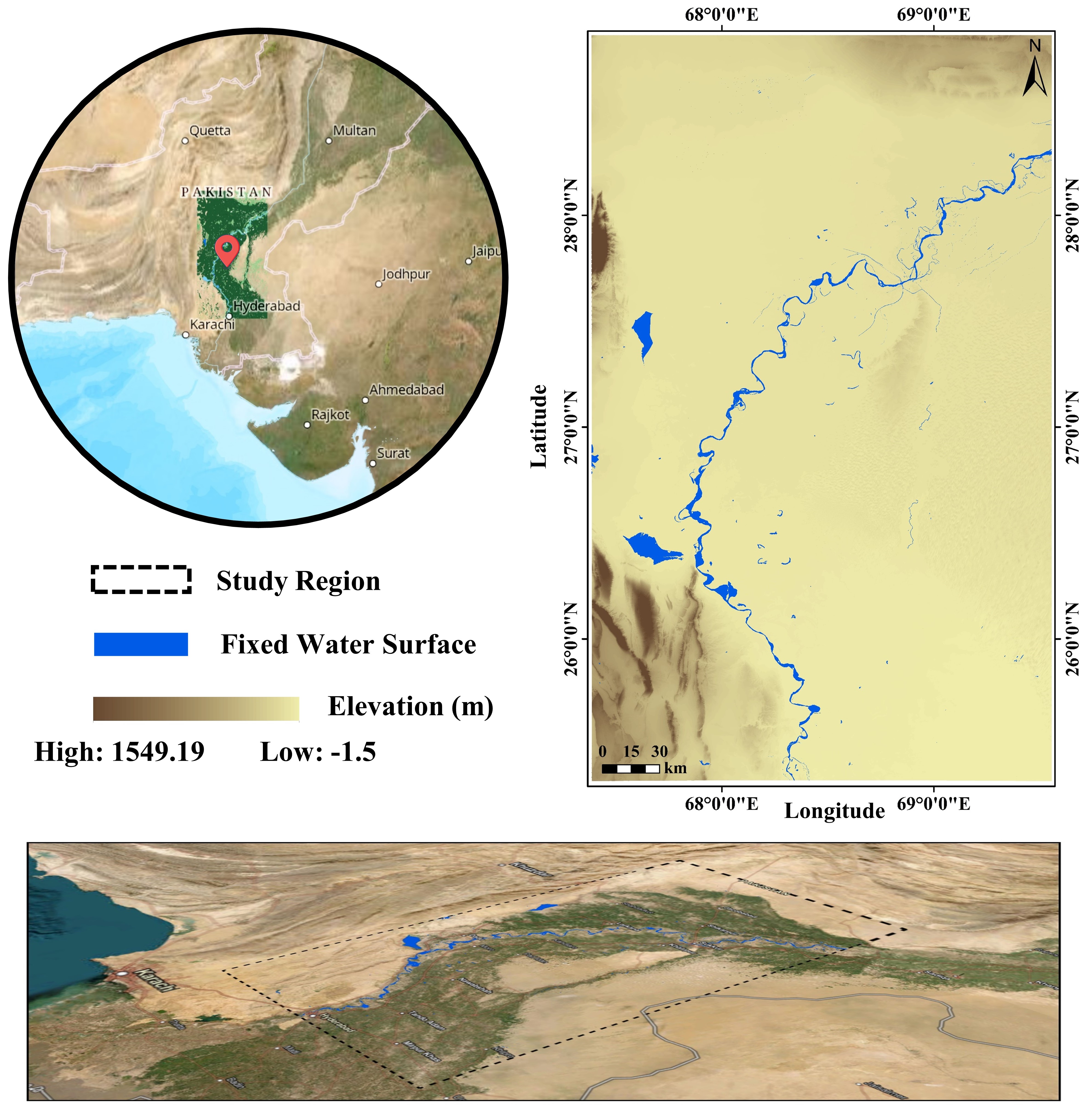}}
	\caption{Location of the study area and elevation information.}
	\label{fig:10}
 \vspace{-4mm}
\end{figure}
Flood events are recurrent phenomena in Pakistan, primarily driven by intense summer monsoon rainfall and occasional tropical cyclones. In the summer monsoon season of 2022, Pakistan experienced a devastating flood event.  This flood event impacted approximately one-third of Pakistan's population, resulting in the displacement of around 32 million individuals and causing the loss of 1,486 lives, including 530 children. 
The economic toll of this disaster has been estimated at exceeding \$30 billion~\citep{Bhutto2022}. Beyond the immediate consequences, the widespread destruction of agricultural fields has raised concerns of potential famine, and there is a looming threat of disease outbreaks in temporary shelters~\citep{nanditha2023pakistan}.

The study area encompasses the regions in Pakistan most severely affected by the flood, spanning the southern provinces of Punjab, Sindh, and Balochistan, covering a total land area of 85,616.5 square kilometers. The Indus River basin, a critical drainage system, plays a pivotal role in this study area's hydrology. Fig.~\ref{fig:10} depicts the study area's location, showcasing a visual representation of its DEM and the river network.

\subsection{Data Requirements for FloodCast}
Data as required by the proposed FloodCast include topographical data, land cover maps, and real-time gridded rainfall data. 
Specifically, a high-resolution (30 m) FABDEM~\citep{hawker202230} from COPDEM30  is required for flood simulation. Considering the constraints imposed by GPU memory limitations and our primary goal of conducting a thorough evaluation of FloodCast's effectiveness and its capacity to maintain resolution invariance, a bilinear interpolation technique is employed to downsample the DEM by factors of 16 (resulting in a spatial resolution of 480m) and 12 (resulting in a spatial resolution of 360m). This process generates both coarse and fine grids, which are used for training and assessing the transferability of the sequence-to-sequence GeoPINS model, respectively. 

Land cover information is useful for estimating and adjusting friction (Manning coefficient) in FloodCast. Land cover information in the study area can be subtracted from a publicly available Globa-Land30 dataset~\citep{chen2015global} developed by the Ministry of Natural Resources of China, which is shown in Fig.~\ref{fig:15} (a). It is a parcel‐based land cover map created by classifying satellite data into 8 classes, available at a spatial resolution of up to 30 m for the study area. Cultivated land is the predominant land cover type (57.05\%) in the Pakistan study area, and urban areas only account for 0.8\% of the total study area.

The rainfall data is a grid‐based data set at $0.1^{\circ} \times 0.1^{\circ}$ spatial resolution and half-hourly temporal resolution from GPM-IMERG. Utilizing the proposed real-time rainfall processing and analysis tool, rainfall data with a temporal resolution of 5 minutes (300 s) and a spatial resolution of $480 m \times 480 m$, as well as data with a temporal resolution of 30 seconds and a spatial resolution of $480 m \times 480 m$ are obtained for the temporal downscaling experiment of FloodCast. Furthermore, rainfall data with a temporal resolution of 5 minutes (300 s) and a spatial resolution of $480 m \times 480 m$, along with rainfall data featuring a temporal resolution of 5 minutes (300 s) and a spatial resolution of $360 m \times 360 m$, are acquired for the spatial downscaling experiment of FloodCast.

Hydrological stations play a pivotal role in furnishing essential data pertaining to inflow boundaries necessary for flood predictions. Utilizing the inflow records obtained from a limited number of stations situated along the Indus River in Pakistan, as disclosed by the Government of Pakistan, and through a comparative analysis of the inflow boundaries within the study area against the nearest hydrological station record data, we have computed the daily discharges for inflow boundary from August 18th to August 31st, as presented in Table~\ref{Tab:Table66}.

\begin{table*}[htp!]
	\caption{Daily discharges for inflow boundary from August 18th to August 31th. }
	\centering
	\resizebox{0.9\textwidth}{!}
	{
	\begin{tabular}{cccccccc}
		\hline
		Dates               & 18-Aug & 19-Aug & 20-Aug & 21-Aug & 22-Aug & 23-Aug & 24-Aug \\
		Discharge at inflow ($m^{3}/s$) & 9345   & 9798   & 10251  & 11667  & 13677  & 15914  & 15489  \\ \hline
		Dates               & 25-Aug & 26-Aug & 27-Aug & 28-Aug & 29-Aug & 30-Aug & 31-Aug \\
		Discharge at inflow ($m^{3}/s$) & 14272  & 13875  & 13875  & 13734  & 14187  & 14527  & 14696  \\ \hline
\end{tabular}}
	\label{Tab:Table66}
\end{table*}

\subsection{Benchmark Datasets for Realistic Flood Simulation and Forecasting}
\begin{figure}[!htp]
	\centering
	{\includegraphics[width = .95\textwidth]{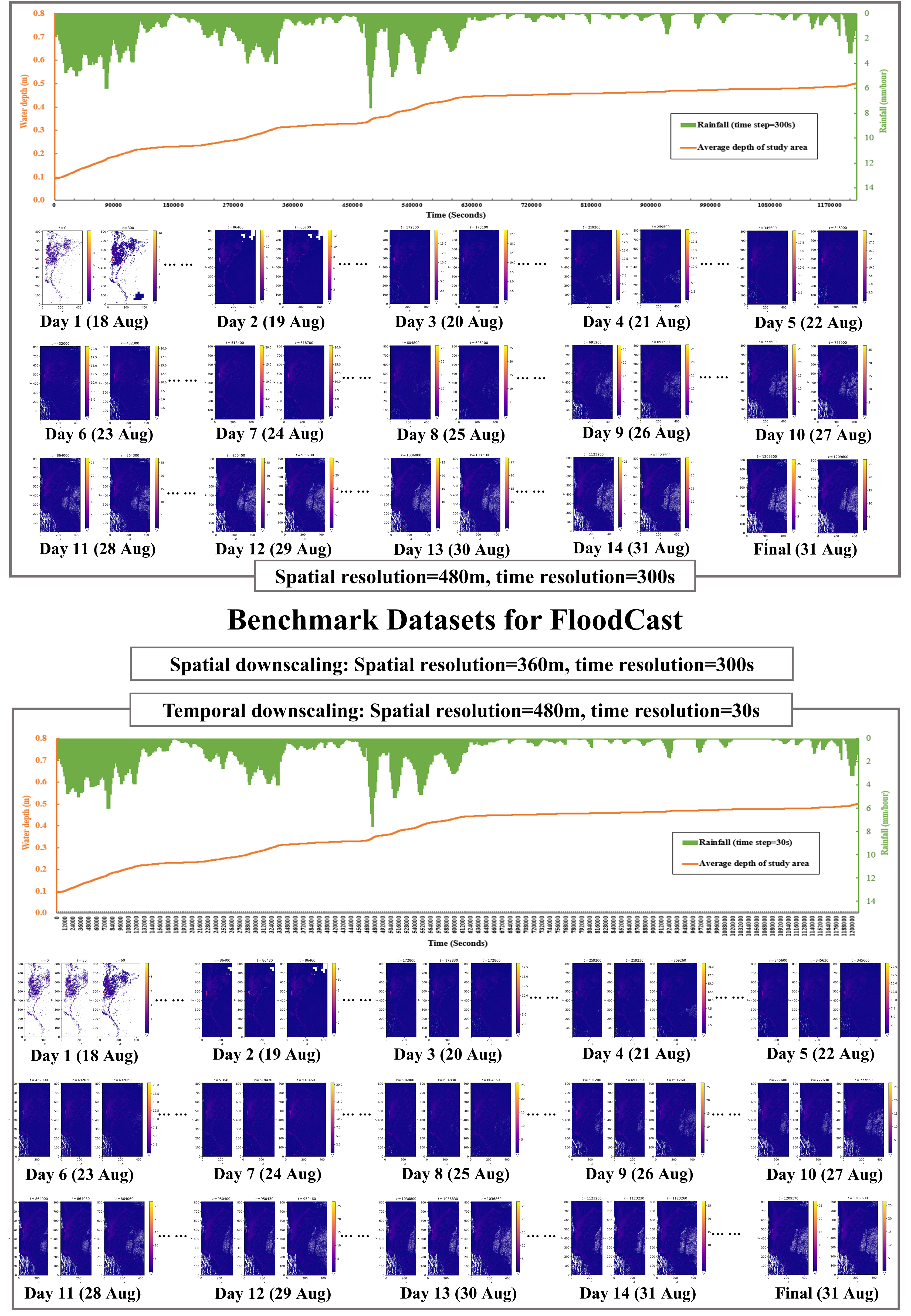}}
	\vspace{-2mm}
	\caption{Benchmark Datasets for FloodCast.}
	\label{fig:16}
	\vspace{-4mm}
\end{figure}
In this section, a  benchmark dataset is established to evaluate different flood prediction and simulation methods. 
We employ a conventional method, as documented in the references~\citep{de2012improving,de2013applicability}, to discretize Eq.~\ref{eq44} from its continuous domain to a discrete domain. This discretization process utilizes a FD scheme applied to a staggered grid. It is worth noting that the FD scheme is a widely accepted numerical solution technique for simulating flood scenarios and is an integral component of the LISFLOOD-FP~\citep{van2010lisflood}, encompassing various forms of SWEs.

The inputs to a hydraulic simulation include an elevation map, initial conditions,  boundary conditions, and the rainfall conditions in the Pakistan study region. Specifically, given the predominant focus of this benchmark dataset on flood inundation extent and inundation depth, the initial conditions mainly consider the water depths obtained from SAR (Fig.~\ref{fig:15} (b)). Regarding boundary conditions, our main consideration is the inflow boundary of the Indus River, which spans the study area. The determination of inflow at the upstream extremity of the river is derived from comprehensive hydrological datasets.  We meticulously specify the inflow width as spanning 10 pixels, precisely centered on the lowest point of the river.   Rainfall is input as a spatial grid, and Manning coefficients are selected based on land cover (Fig.~\ref{fig:15} (a)), and FLO-2D references~\citep{o2011flo}. Furthermore, the discrete implementation described in~\citep{de2013applicability} uses two parameters - a weighting factor $\theta$ that adjusts the amount of artificial diffusion, and a coefficient $0<\alpha \leq 1$ that is used as a factor by which we multiply the time step. We use the proposed values in~\citep{de2013applicability}, namely $\theta=0.7, \alpha=0.7$. 

We implement the numerical solution using Python for a 14-day period, from August 18 to August 31, totaling 1,209,600 seconds, within the Pakistan study area. We obtain results at two spatial resolutions: 480m $\times$ 480m (coarse grids) and 360m $\times$ 360m (refined grids). Temporal resolution adheres to the Courant-Friedrichs-Lewy (CFL) condition for hyperbolic system stability.
Due to the fine temporal resolution (less than 10s), the 14-day simulations at 480m $\times$ 480m resolution run on an NVIDIA A6000 GPU in about one week, while those at 360m $\times$ 360m resolution take approximately 1.5 weeks. 
Finally, our benchmark dataset for FloodCast can be generated, as illustrated in Fig.~\ref{fig:16}. This benchmark dataset comprises three subsets, including a temporal resolution of 300s and a spatial resolution of 480m for model training and validation, temporal downscaling (temporal resolution 30s, spatial resolution 480m) and spatial downscaling (spatial resolution 360m, temporal resolution 300s) for the spatiotemporal downscaling transferability. The Fig.~\ref{fig:16} curve shows changes in the average water depth in the study area in response to rainfall. It is evident that the continuous rainfall from August 18 to August 25 leads to a sustained increase in regional water depths. Notably, the depths increase more rapidly before August 25 when rainfall is relatively high. Subsequently, as rainfall diminishes after August 25, the average water depths exhibit a slower rate of increase. These observed trends serve as preliminary confirmation of the reliability of our benchmark dataset.

\section{Experiments and Results of Multi-satellite Observations}
 In this section, the accuracy and efficiency of the proposed multi-satellite observations in FloodCast are evaluated. 
\subsection{Validation of unsupervised change detection}
 \begin{figure}[!htp]
 	\centering
 	{\includegraphics[width = .85\textwidth]{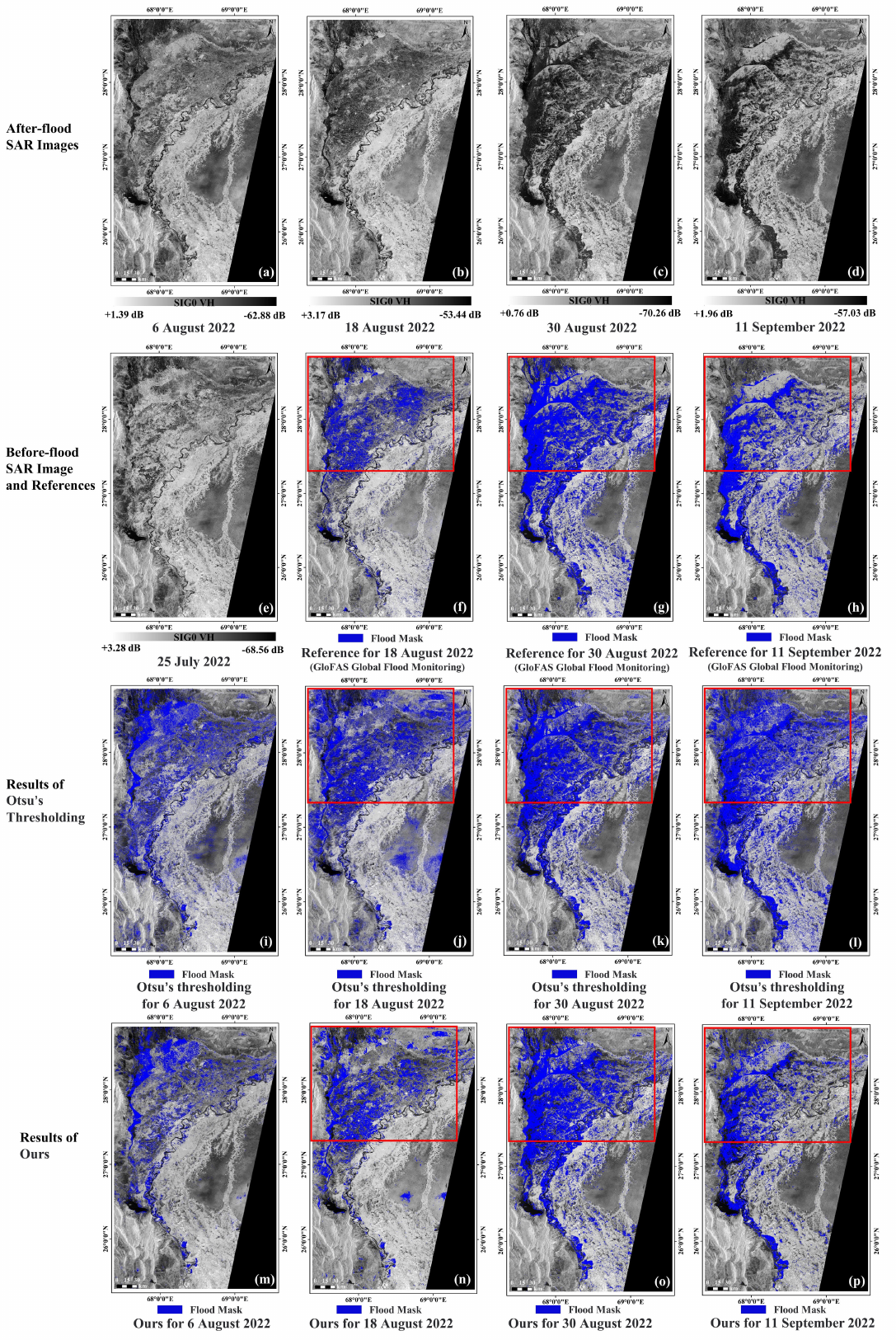}}
 	\vspace{-2mm}
 	\caption{Validation of real-time UCD.  The red box represents the focus regions.}
 	\label{fig:13}
 	\vspace{-4mm}
 \end{figure}
 We report the results of UCD in Fig.~\ref{fig:13}. Four time-series post-flood SAR images (i.e. 6th August 2022, 18th August 2022, 30th August 2022, 11th September 2022) are acquired in Fig.~\ref{fig:13} (a)-(d), respectively. The before-flood SAR image is obtained on 25th July 2022 (Fig.~\ref{fig:13} (e)). 
 It is evident that SAR images display significantly reduced backscattering values (indicated by darker regions in Fig.~\ref{fig:13} (a)-(e)) over all water bodies, enhancing their distinguishability for water change detection.  When the flood regions have been identified, it is necessary to select adequate reference results for verifying the results of the proposed method. Accordingly, the Sentinel-1-based flood mapping results for the period of August 10 to September 11, 2022, provided by the Global Flood Awareness System (GloFAS) Global Flood Monitoring~\citep{roth2022sentinel}, have been designated as the reference results. These reference results are depicted in Fig.~\ref{fig:13} (f)-(h), respectively. We compare our proposed scaled Otsu's thresholding with traditional Otsu's thresholding. Comparing the time series flood mapping results generated by our proposed method (Fig.~\ref{fig:13} (m)-(p)) with that of Otsu's thresholding (Fig.~\ref{fig:13} (i)-(l)), we observe that Otsu's thresholding tends to over-identify flood areas, resulting in lower precision for flood regions. For instance,  comparing the flood mapping on 11th September 2022 (Fig.~\ref{fig:13} (h), Fig.~\ref{fig:13} (l), and Fig.~\ref{fig:13} (p)), Otsu's thresholding easily misclassifies non-flooded regions as floods. In contrast, our proposed method achieves high detection accuracy and finer results due to the adaptive thresholding.  Notably, based on the GEE platform, our proposed method produces results within 10 seconds for each flood mapping, allowing us to obtain flood inundation results for the study area from August 6 to September 11, 2022, within 1 minute. 
Based on the time-series flood masks across four distinct time periods (Fig.~\ref{fig:13} (m)-(p)), it is evident that the flood surface area consistently expanded from August 6, reaching its peak on August 30 (a flooded area of 11,063$km^{2}$). A notable change occurred between August 18 and August 30, marked by significant increases in flood coverage in both northern and southern regions. Consequently, we focus on this timeframe, starting from August 18, and employ earth observation data and the proposed sequence-to-sequence GeoPINS model to predict the high-resolution spatiotemporal evolution pattern and maximum flood coverage during this period. 

To accurately evaluate the effectiveness of the proposed UCD method, we conducted a quantitative comparison of the flood mask results for August 18 and August 30, as presented in Table~\ref{Tab:Table4}.  The Kappa coefficient (Kappa) and overall accuracy (OA) are used to evaluate overall prediction accuracy.  Precision (the ability of a model to not label a true negative observation as positive) and recall (the model's capability to find positive observations) are utilized to quantitatively evaluate the performance with respect to flood regions and non-flood regions. The results clearly demonstrate that our method outperforms Otsu's thresholding-based approach significantly. In particular, there is a notable improvement in Kappa, with increases of 9.14\% and 1.75\% observed in comparison to Otsu's thresholding on August 18 and August 30, respectively.
 \begin{table*}[htp!]
	\caption{Accuracy assessment of real-time UCD. The values in \textbf{bold} are the best. }
	\centering
	\resizebox{0.9\textwidth}{!}
	{
	\begin{tabular}{cccccccc}
		\hline
		&                           &                         &                      & \multicolumn{2}{c}{Flood Region} & \multicolumn{2}{c}{Non-flood region} \\ \cline{5-8} 
		\multirow{-2}{*}{Dates}                            & \multirow{-2}{*}{Methods} & \multirow{-2}{*}{Kappa} & \multirow{-2}{*}{OA} & Recall          & Precision      & Recall            & Precision        \\ \hline
		& Otsu’s Thresholding       & 28.01                   & 85.00                & \textbf{57.99}  & 25.08          & 87.03             & \textbf{96.51}   \\ \cline{2-8} 
		\multirow{-2}{*}{18-Aug-22}                        & Ours                      & \textbf{37.15}          & \textbf{90.59}       & 49.22           & \textbf{36.87} & \textbf{93.69}    & 96.10            \\ \hline
		{\color[HTML]{2A2B2E} }                            & Otsu’s Thresholding       & 59.71                   & 88.03                & 65.94           & 68.13          & 93.03             & 92.36            \\ \cline{2-8} 
		\multirow{-2}{*}{{\color[HTML]{2A2B2E} 30-Aug-22}} & Ours                      & \textbf{61.46}          & \textbf{88.60}       & \textbf{66.97}  & \textbf{69.92} & \textbf{93.49}    & \textbf{92.60}   \\ \hline
\end{tabular}}
\label{Tab:Table4}
\end{table*}

\begin{figure}[!htp]
	\centering
	{\includegraphics[width = .85\textwidth]{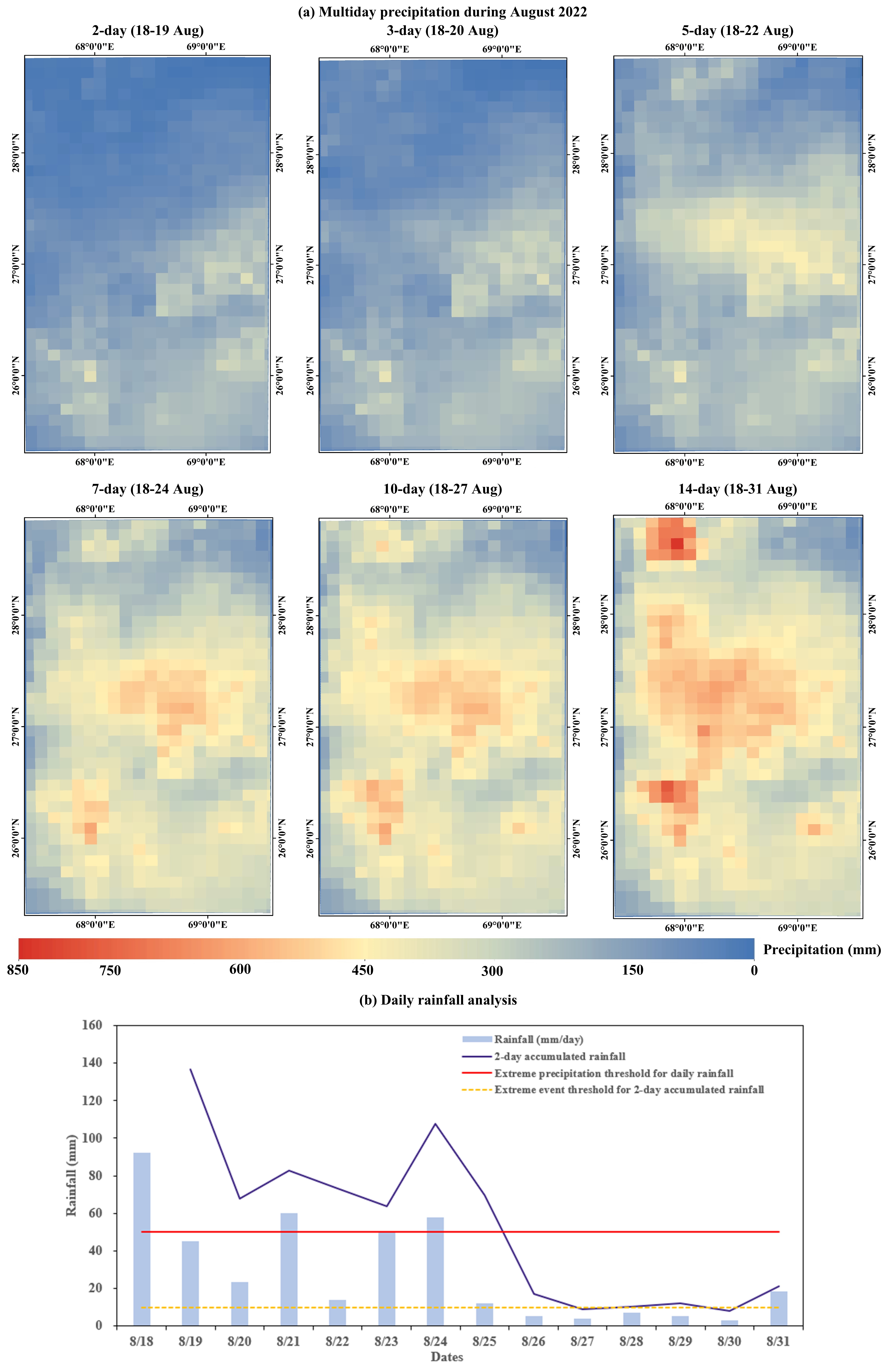}}
	\vspace{-2mm}
	\caption{Rainfall analysis during the 2022 flood in Pakistan.}
	\label{fig:14}
	\vspace{-4mm}
\end{figure}
\subsection{Real-time rainfall analysis}
Next, we examine the rainfall intensity and spatiotemporal distribution in Pakistan from August 18 to August 30  to make a decision regarding the initiation of the FloodCast model. 
We analyze the spatial distribution of precipitation at different durations during August 2022 to identify the region that received heavy precipitation (Fig.~\ref{fig:14} (a)). We demarcate the extreme precipitation region using the spatial distribution of 14-day precipitation. Several regions (roughly 37,844$km^{2}$) experienced 14-day accumulated precipitation greater than 400mm.

To conduct a more comprehensive evaluation of the magnitude of rainfall, we defined different thresholds for analysis, shown in Fig.~\ref{fig:14} (b). First, we defined extreme events as days exceeding 10 mm of 2-day accumulated rainfall over Pakistan~\citep{webster2011were}. 
The selected threshold is considerably smaller than the maximum daily rainfall observations at individual stations because of the greater averaging area.  By comparing the 2-day accumulated rainfall (blue line in Fig.~\ref{fig:14} (b)) and extreme event threshold for 2-day accumulated rainfall (yellow dotted line in Fig.~\ref{fig:14} (b)) from August 18 to August 31, it is evident that the cumulative rainfall exceeded the designated threshold for the majority of the time (i.e. seven consecutive days from August 19 to August 26, August 28, August 29, and August 31). Notably, the highest two-day cumulative rainfall was recorded on August 19, totaling 136.81mm. Furthermore, we analyzed daily rainfall over the study period. The frequency of extreme precipitation events for the summer season is defined based on the thresholds of daily precipitation as 50 mm/day~\citep{ikram2016past}. It can be seen from Fig.~\ref{fig:14} (b) that some days (August 18, August 21, August 23, August 24) experienced precipitation greater than the extreme precipitation threshold for daily rainfall (red line in Fig.~\ref{fig:14} (b)). The highest daily extreme precipitation (92.11mm) occurred on 18 August. Thus, accurate flood prediction with a high spatiotemporal resolution is essential for the high frequency of extreme rainfall periods.

\begin{figure}[!htp]
	\centering
	{\includegraphics[width = .95\textwidth]{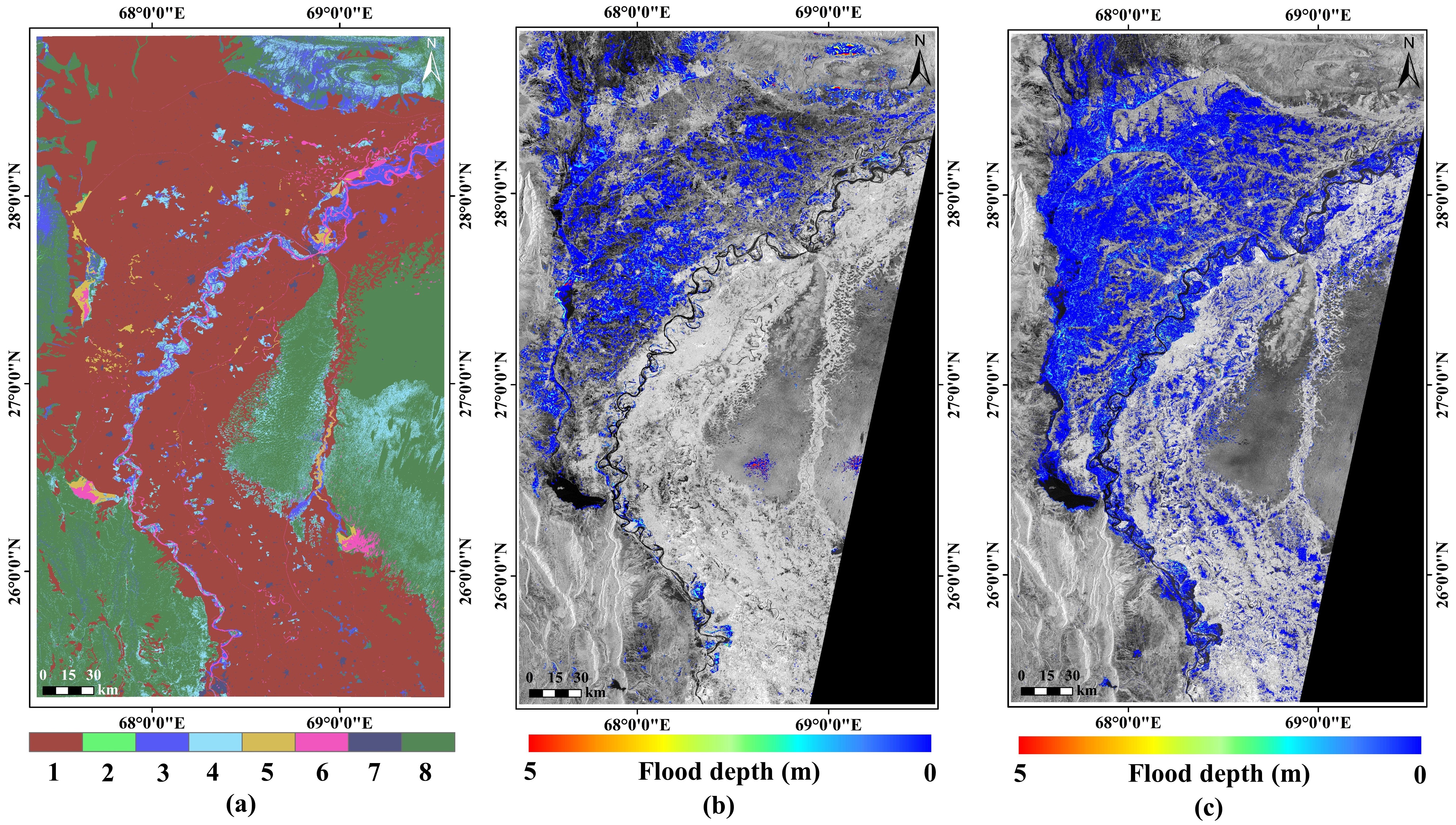}}
	\vspace{-2mm}
	\caption{(a) Land use and land cover. 1: Cultivated land, 2: Forest, 3: Grass land, 4: Shrubland, 5: Wetland, 6: Water body, 7: Artificial surfaces, 8: Bare land; (b) Flood depth extracted from UCD on 18 August 2022; (c) Flood depth extracted from GloFAS on 30 August 2022.}
	\label{fig:15}
	\vspace{-4mm}
\end{figure}

\subsection{Flood depth based on Earth observation}
To initiate and validate the hydrodynamic model, the flood inundation depths on August 18 and August 30, 2022, are extracted using FABDEM and the SAR-based flood extent, as shown in Fig.~\ref{fig:15} (b) and Fig.~\ref{fig:15} (c), respectively. Specifically, with the explicit objective of enhancing the utilization of UCD results within the FloodCast to uphold the essential real-time flood prediction, on August 18, the flood depth (Fig.~\ref{fig:15} (b)) ascertained through UCD, and the water depth in fixed water areas (such as lakes and rivers), will serve as the initial conditions for starting the hydrodynamic model. The determination of water depth in these fixed water bodies is accomplished through the utilization of FwDET in conjunction with the water surface of the study area, extracted from OpenStreetMap~\citep{yamazaki2017high}, and FABDEM. In addition,  the average water depth is 0.0936m in the study area on August 18. Furthermore, in pursuit of minimizing uncertainties within the model validation process,
the flood depth extracted from the Global Flood Awareness System (GloFAS) on August 30 will be utilized to verify the results of hydrodynamic model. As depicted in Fig.~\ref{fig:15} (c),  it is evident that the flood depth is notably elevated in proximity to the riverbanks and concentrated flood-prone regions, such as Larkana, Sukkur, Shikarpur, Shahdadkot, and other regions. The average water depth within the study area on August 30 is 0.5471m. There might be uncertainties in the flood depth within the study region due to the inherent inaccuracies of FABDEM and estimated inundation extent.

\section{Experiments and Results of Sequence-to-sequence GeoPINS}
In this section, a rigorous evaluation is performed to assess the precision and efficiency of the proposed sequence-to-sequence GeoPINS and zero-shot super-resolution features within the FloodCast framework.

\subsection{Validation of Sequence-to-sequence GeoPINS}
To evaluate the performance of the unsupervised hydrodynamic model, sequence-to-sequence GeoPINS, for large-scale flood simulation and forecast,  we conduct a rapid simulation of the study area at a coarse spatiotemporal resolution (spatial resolutions = 480m $\times$ 480m and temporal resolution = 300s) over 14 consecutive days, from August 18th to August 31st. The time length of each input domain $\Delta T$ is equal to 16 in the sequence-to-sequence GeoPINS, resulting in a total of 268 consecutive sequences. To mitigate the accumulation of errors in initial conditions during the sequence-to-sequence learning, the initial conditions of the first 200 time series from the numerical simulation are utilized as in-situ observational data for the proposed sequence-to-sequence GeoPINS. The entire 14-day simulation, conducted on an NVIDIA A6000, requires 2.419 days to complete. Furthermore, through comprehensive comparisons with benchmark datasets for FloodCast and Earth observation results, we have substantiated the efficacy of our model in accurately predicting flood inundation extents and depths.
\begin{figure}[!h]
	\centering
	{\includegraphics[width = .9\textwidth]{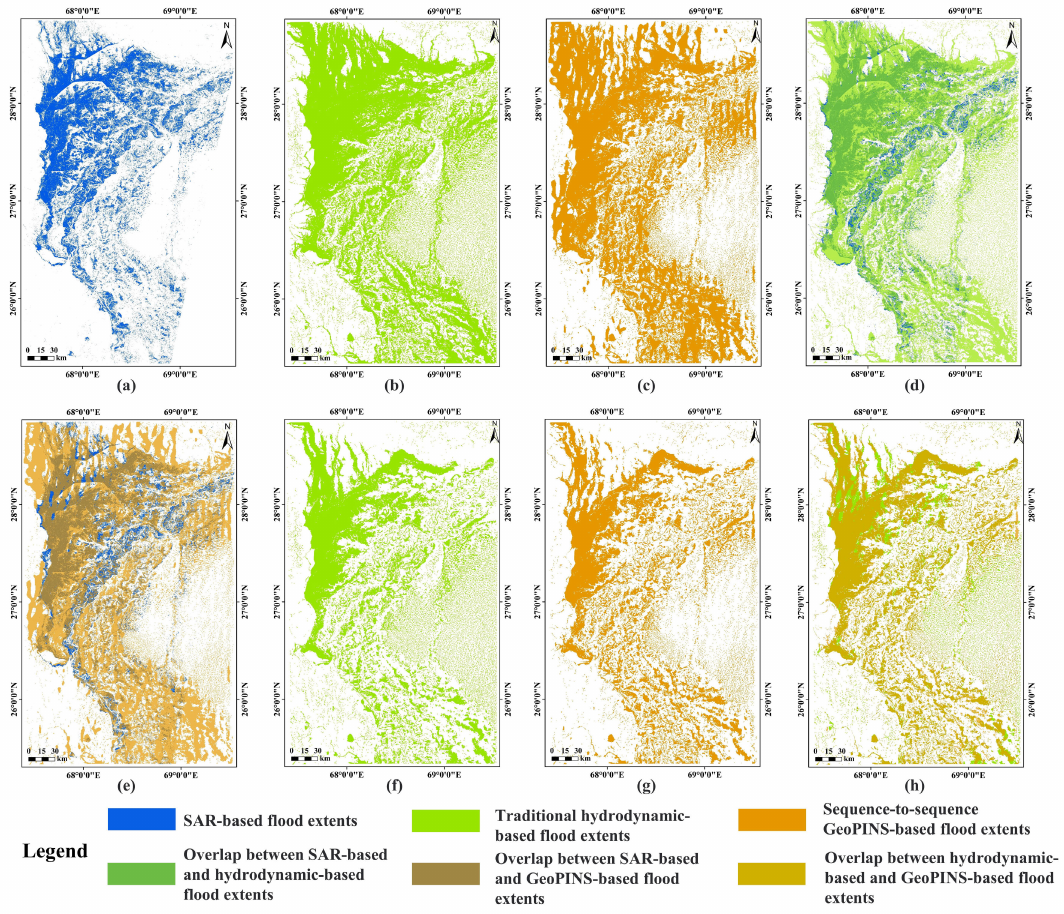}}
 	\caption{Flood extent results on 30 August 2022. (a) SAR-based flood extent on 30 August 2022. (b) Traditional hydrodynamic method (benchmark)-based flood extent (flood depth $\geq$ 0.1m). (c) Sequence-to-sequence GeoPINS-based flood extent (flood depth $\geq$ 0.1m). (d) Comparison between SAR-based flood extents and traditional hydrodynamic method-based flood extent (flood depth $\geq$ 0.1m). (e)  Comparison between SAR-based and sequence-to-sequence GeoPINS-based flood extents. (f) Traditional hydrodynamic method (benchmark)-based flood extent (flood depth $\geq$ 0.5m). (g) Sequence-to-sequence GeoPINS-based flood extent (flood depth $\geq$ 0.5m). (h) Comparison between traditional hydrodynamic method-based and sequence-to-sequence GeoPINS-based flood extents (flood depth $\geq$ 0.5m).}
	\label{fig:17}
	\vspace{-4mm}
\end{figure}

\textbf{Validation of flood inundation extents.} Fig.~\ref{fig:17} presents the flood inundation extents on August 30, obtained by different methods 
and their comparative analysis. The traditional hydrodynamic-based inundation extent (Fig.~\ref{fig:17} (b)) and sequence-to-sequence GeoPINS-based flood extent (Fig.~\ref{fig:17} (c)) are obtained when the flood depth exceeds 0.1m. Evidently, the overall inundation boundaries of the two methods exhibit a reasonable degree of consistency; however, notable disparities become apparent in the upper portion of the study area.  
To further validate these two results, we conducted overlapping assessments with SAR-based flood extent (Fig.~\ref{fig:17} (a)), as presented in Fig.~\ref{fig:17} (d) and Fig.~\ref{fig:17} (e).  Remarkably, both the hydrodynamic-based and sequence-to-sequence GeoPINS-based inundation extents encompass a significant portion of the SAR area, aligning their overall flood extents. However, hydrodynamic results continue to exhibit superior coverage in the upper study area when compared to the results of sequence-to-sequence GeoPINS. 
Furthermore,  through a comparative assessment of SAR-based flood extent and two hydrodynamic-based flood extents (Fig.~\ref{fig:17} (d) and Fig.~\ref{fig:17} (e)),  in conjunction with the land use and land cover data (Fig.~\ref{fig:15} (a)),  it has been discerned that the areas exhibiting non-congruence with the SAR-based flood extents are, indeed, characterized as cultivated land. This incongruity may be attributed to the circumstance in which the water depth fails to exceed the height of the crops, resulting in minimal alteration in the backscattering characteristics within the SAR image and, consequently, the absence of flood detection within this region of the SAR imagery.
For a detailed comparison at higher water depths, we generated inundation extents exceeding 0.5m and created overlay maps, as presented in Fig.~\ref{fig:17} (f), Fig.~\ref{fig:17} (g), and Fig.~\ref{fig:17} (h), respectively. Upon thorough examination, we found a notable congruence between the two methods in both finer details and overall extent. 

\begin{figure}[!htp]
	\centering
	{\includegraphics[width = 0.9\textwidth]{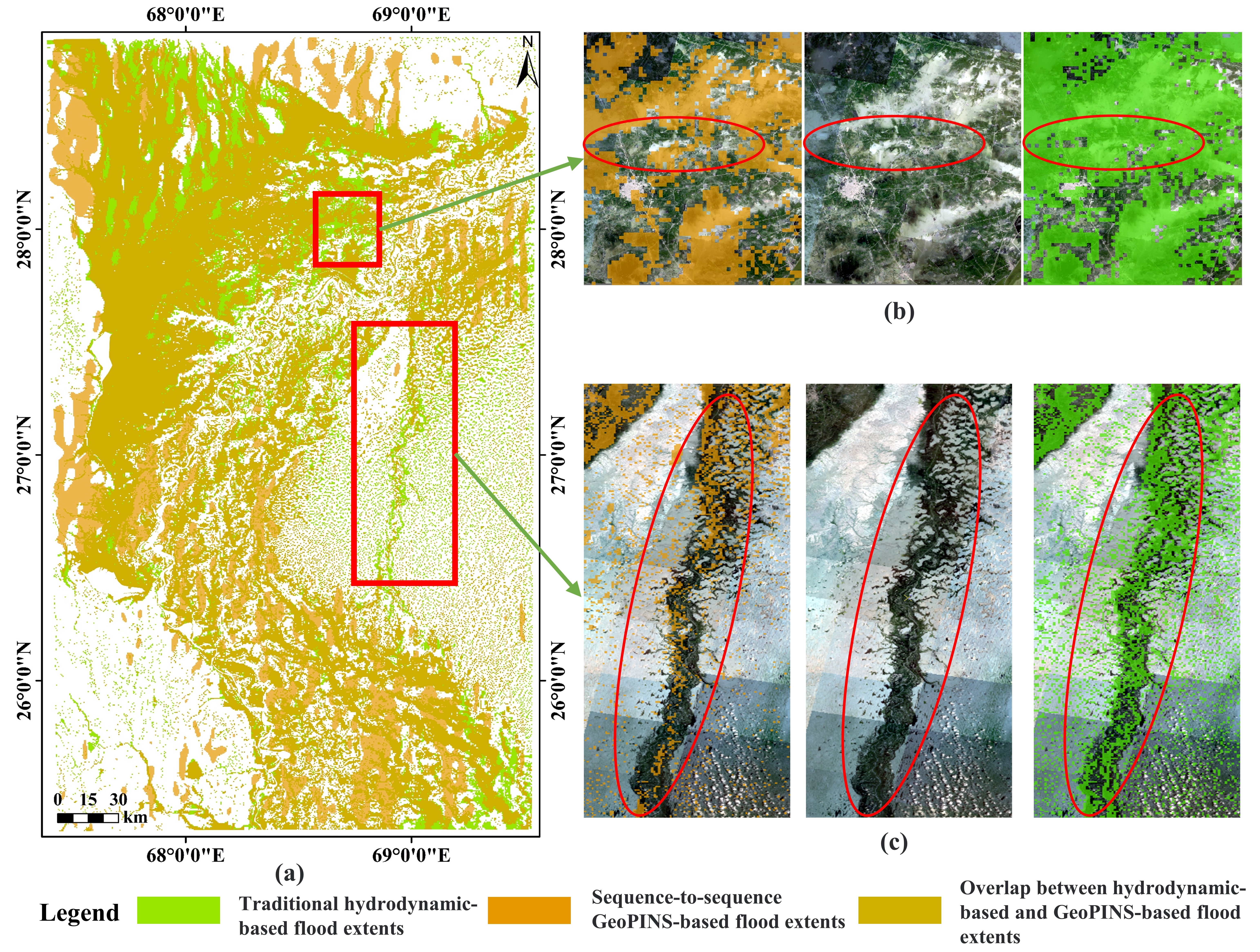}}
	\vspace{-2mm}
	\caption{Comparison between traditional hydrodynamic method (benchmark)-based and sequence-to-sequence GeoPINS-based flood extents (flood depth $\geq$ 0.1m). 
 (b) and  (c) depict high-resolution optical data comparison depicting the same timeframe on August 30, featuring the sequence-to-sequence GeoPINS-based flood extent, the original remote sensing image, and traditional hydrodynamics-based flood extent, displayed from left to right.
 The red box represents the focus regions.
 }
	\label{fig:18}
\end{figure}

To further view the dissimilarities in submersion extent details between hydrodynamics and sequence-to-sequence GeoPINS and to mitigate uncertainties associated with SAR-based results,  we acquire high-resolution optical remote sensing images from the Planetscope satellite for August 30, with minimal cloud cover.   We selectively focus on areas displaying distinct submersion extent disparities between hydrodynamics and sequence-to-sequence GeoPINS (Fig.~\ref{fig:18} (a)). Concurrently, we obtained high-resolution optical data depicting the same timeframe on August 30, as exemplified in Fig.~\ref{fig:18} (b) and Fig.~\ref{fig:18} (c).  Fig.~\ref{fig:18} (b) and Fig.~\ref{fig:18} (c) display the sequence-to-sequence GeoPINS-based flood extent,  the original remote sensing image, and traditional hydrodynamics-based flood extent
from left to right, respectively.
This comparison highlights that hydrodynamics occasionally misidentifies vegetation as flooded areas (over-recognition), whereas sequence-to-sequence  GeoPINS offers more precise forecasting of details.

In summary, the visualization results indicate that traditional hydrodynamics-based flood extents align better with SAR-based results in the peripheral regions, while sequence-to-sequence GeoPINS-based flood extents excel in detail accuracy within inundated areas. Both methods demonstrate excellent agreement under high water levels.

\begin{figure}[!htp]
	\centering
	{\includegraphics[width = 0.9\textwidth]{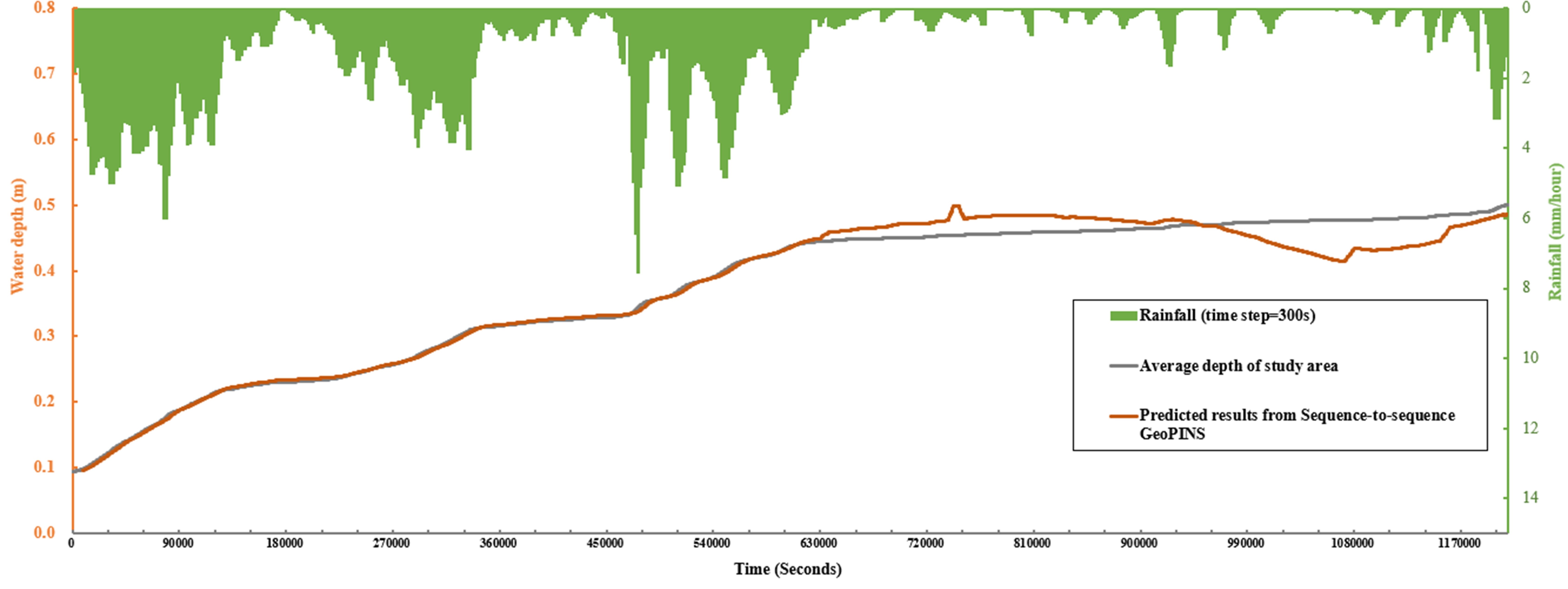}}
	\vspace{-2mm}
	\caption{Comparison of the average depth of the study area calculated using the traditional hydrodynamics method and the average depth computed by sequence-to-sequence GeoPINS over a 14-day period of rainfall (time step=300s).}
	\label{fig:19}
	\vspace{-4mm}
\end{figure}
\textbf{Validation of flood depths.}  Fig.~\ref{fig:19} presents a comparative analysis of the average depth in the study area, as determined by the traditional hydrodynamic modeling and sequence-to-sequence GeoPINS, over a 14-day period of varying rainfall conditions (time step = 300s). The general trend between the two methods aligns consistently. Specifically,  before August 25, heavy rainfall led to a significant increase in regional average depth, followed by a slower rate of change under reduced rainfall after August 25. Notably, both the traditional hydrodynamics and sequence-to-sequence GeoPINS are generally in agreement for the first 63,000 seconds. However, disparities emerge thereafter due to the cumulative error associated with model training,  where initial values are provided for the first 200 time series by using in-site observations, but later time steps rely on the previous step's predicted values as initial conditions. Further scrutiny of sequence-to-sequence GeoPINS' average depth change curve indicates that its predictions are reasonable. Concretely, between 900,000 seconds and 1,080,000 seconds, the continuous decrease in rainfall results in a decline in average depth, which is followed by a significant increase with the resumption of rainfall after 1,080,000 seconds.

\begin{figure}[!htp]
	\centering
	{\includegraphics[width = 1.0\textwidth]{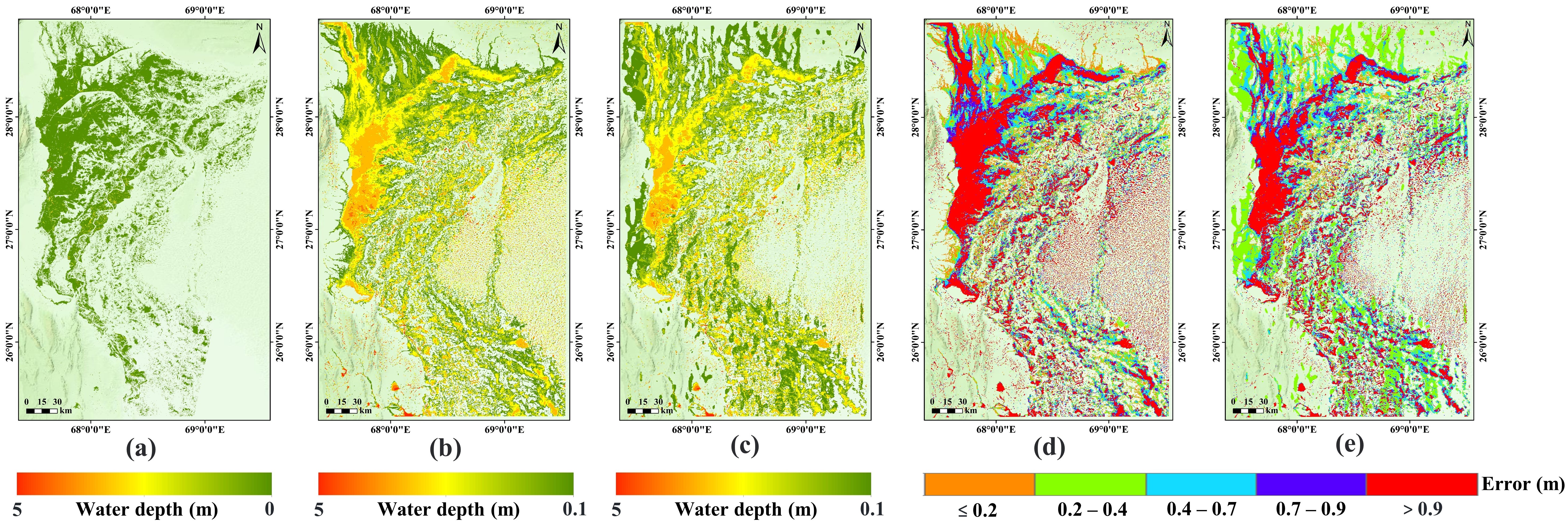}}
	\vspace{-2mm}
	\caption{Flood depths on 30 August 2022. (a) SAR-based flood depth. (b) Traditional hydrodynamic method (benchmark)-based flood depth (flood depth $\geq$ 0.1m). (c) Sequence-to-sequence GeoPINS-based flood depth (flood depth $\geq$ 0.1m). (d) Spatial distribution of the errors between SAR-based and traditional hydrodynamic method-based flood depths. (e)  Spatial distribution of the errors between SAR-based and sequence-to-sequence GeoPINS-based flood extents. }
	\label{fig:20}
	\vspace{-4mm}
\end{figure}

To further examine and compare the spatiotemporal variations in flood depth between the traditional hydrodynamics and sequence-to-sequence GeoPINS, we compare flood depth maps for August 30 and SAR-based flood depth maps for the same date (Fig.~\ref{fig:20}). Fig.~\ref{fig:20} (a) to Fig.~\ref{fig:20} (c) depict flood depth results derived from SAR, traditional hydrodynamics, and sequence-to-sequence GeoPINS, respectively, with most flood depths falling within the 0.1-3.0m range. Additionally, employing  SAR-based flood depth for comparative analysis, we assess the spatial error distribution of flood depth for the traditional hydrodynamics and sequence-to-sequence GeoPINS, as illustrated in Fig.~\ref{fig:20}  (d) and Fig.~\ref{fig:20} (e). It is evident that sequence-to-sequence GeoPINS' flood depth error map exhibits significantly smaller errors than that of hydrodynamics in areas with depths exceeding 0.9m (red in Fig.~\ref{fig:20}). Further utilizing SAR-based results as a reference, we quantitatively calculate the Mean Absolute Error (MAE) for traditional hydrodynamics and sequence-to-sequence GeoPINS, yielding values of 0.4203m and 0.3615m, respectively. 
MAE, which ranges from [0,$+\infty$), is a metric for measuring the difference between predicted values and observed values, with smaller MAE values indicating higher prediction accuracy.
These quantitative and qualitative findings validate the reliability and effectiveness of our proposed model's prediction results to a significant extent. 

\begin{figure}[!htp]
	\centering
	{\includegraphics[width = 0.75\textwidth]{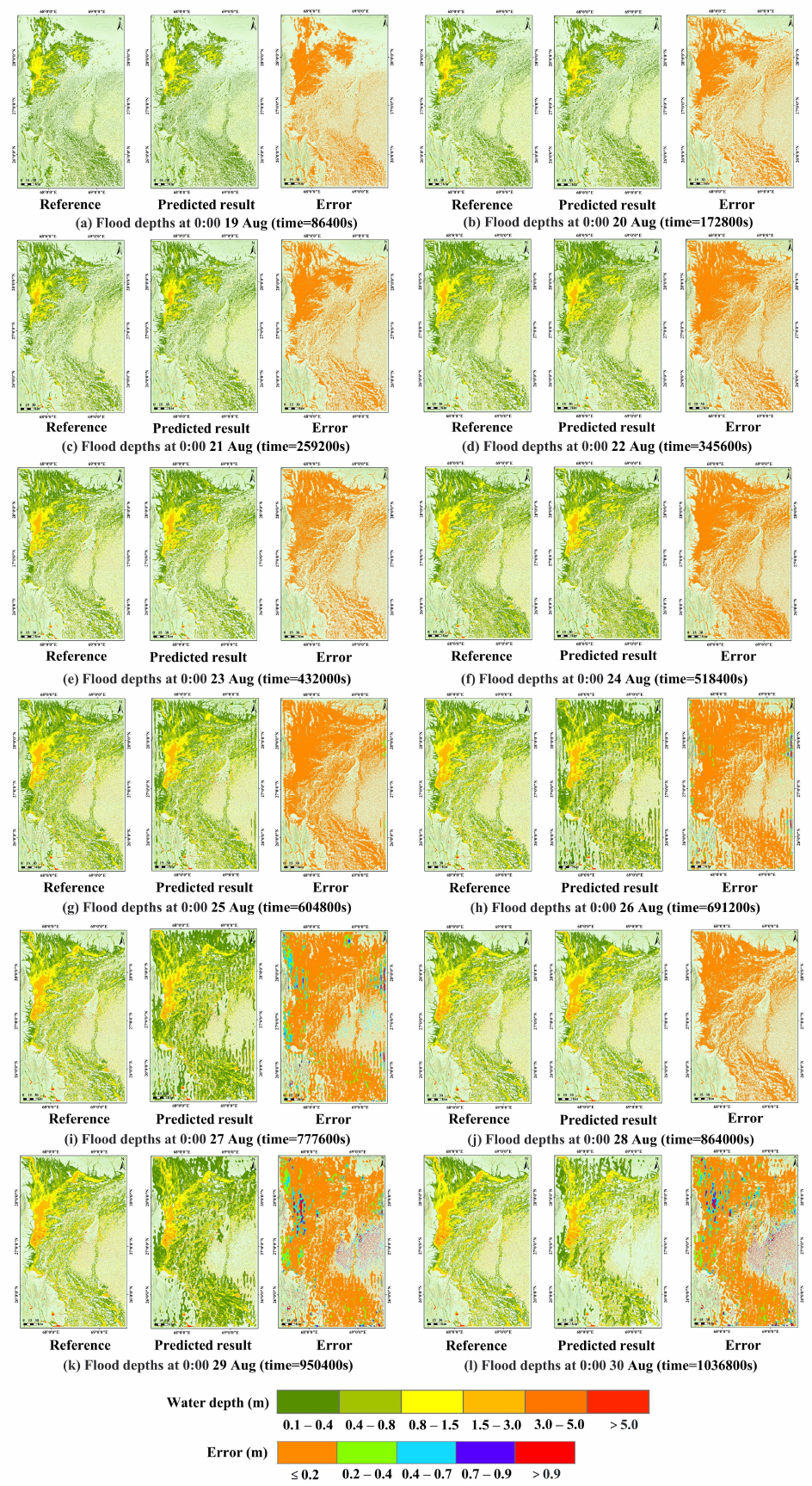}}
	\vspace{-2mm}
	\caption{The flood depths created from the traditional hydrodynamic method and sequence-to-sequence GeoPINS during different stages of the Pakistan flood 2022.}
	\label{fig:21}
	\vspace{-4mm}
\end{figure}

Furthermore, we conduct a spatial error analysis by comparing daily flood spatial distribution maps generated by traditional hydrodynamic modeling and sequence-to-sequence GeoPINS at 0:00 on each day from August 19th to August 30th (Fig.~\ref{fig:21}). Using a 0.1m threshold to distinguish wet and dry cells. It is evident that, from August 19th to August 25th, sequence-to-sequence GeoPINS consistently exhibits error values within 0.2m compared to hydrodynamics' benchmark results. However, from August 25th to August 30th, a few areas in the study region display error heights exceeding 0.2m, although the majority of areas still exhibit errors within the 0.2m range. This once again underscores the effectiveness and accuracy of our proposed model in large-scale flood prediction across both spatial and temporal dimensions.

\begin{table*}[htp!]
	\caption{Accuracy assessment of flood depths. }
	\centering
	\resizebox{1.0\textwidth}{!}
	{\begin{tabular}{c|cccccccccccccc|c}
			\hline
			\textbf{Forecasting Errors}                              & \textbf{18 Aug} & \textbf{19 Aug} & \textbf{20 Aug} & \textbf{21 Aug} & \textbf{22 Aug} & \textbf{23 Aug} & \textbf{24 Aug} & \textbf{25 Aug} & \textbf{26 Aug} & \textbf{27 Aug} & \textbf{28 Aug} & \textbf{29 Aug} & \textbf{30 Aug} & \textbf{31 Aug} & \textbf{Overall Errors} \\ \hline
			\makecell{MAPE (\%) with water pixels \\ (water   depth $\geq$ 0m)}      & 5.70\%                  & 1.46\%                  & 1.03\%                  & 1.01\%                  & 1.23\%                  & 1.79\%                  & 2.57\%                  & 9.08\%                  & 16.37\%                 & 21.17\%                  & 26.63\%                  & 38.27\%                  & 57.51\%                  & 67.69\%                  & 14.93\%                   \\ \hline
			\makecell{MAE (m) with water pixels \\  (water depth $\geq$ 0m)}   & 0.0103                  & 0.0051                  & 0.0045                  & 0.0046                  & 0.0059                  & 0.0087                  & 0.0133                  & 0.0459                  & 0.0824                  & 0.0940                   & 0.1025                   & 0.1329                   & 0.2355                   & 0.3067                   & 0.0610                    \\ \hline
			\makecell{ MAE (m) with water pixels \\  (water depth $\geq$ 0.1m)} & 0.0076                  & 0.0027                  & 0.0022                  & 0.0027                  & 0.0035                  & 0.0053                  & 0.0096                  & 0.0427                  & 0.0805                  & 0.0925                   & 0.1010                   & 0.1317                   & 0.2357                   & 0.3075                   & 0.0589                    \\ \hline
			\makecell{MAE (m) with  water pixels \\ (water depth $\geq$ 0.5m)}  & 0.0060                  & 0.0015                  & 0.0013                  & 0.0016                  & 0.0022                  & 0.0036                  & 0.0064                  & 0.0295                  & 0.0557                  & 0.0647                   & 0.0735                   & 0.1042                   & 0.1941                   & 0.2662                   & 0.0452                    \\ \hline
		\end{tabular}}
	\label{Tab:Table5}
\end{table*}

To further quantify sequence-to-sequence  GeoPINS' performance in reproducing the spatial predictions of traditional hydrodynamic modeling (FD), we employ MAE and Mean Absolute Percentage Error (MAPE) as performance evaluation metrics. 
The 14-day flood depth data is used to calculate the daily average MAE and MAPE under different water depth thresholds (0m, 0.1m, 0.5m) as well as the overall average MAE and MAPE for the entire 14-day period (as summarized in Table~\ref{Tab:Table5}). It is evident that from August 18th to August 31st, both MAE and MAPE values increase with time and under different water depth thresholds. When the water depth threshold is set to 0m, GeoPINS achieves an average MAPE of 14.93\% and an average MAE of 0.0610m for its 14-day water depth predictions. This result reaffirms the effectiveness of our method in long-term spatiotemporal prediction of large-scale floods. Notably, we observe that as the water depth threshold increases (from 0m to 0.5m), the 14-day MAE average gradually decreases, further validating the model's reliability in predicting high water depths.

In summary, we verify the accuracy and effectiveness of our proposed sequence-to-sequence GeoPINS in predicting water depth over large spatial and temporal scales through both qualitative and quantitative analyses. Comparative assessments with SAR-based flood depth indicate that sequence-to-sequence GeoPINS' flood depth predictions exhibit smaller errors when compared to traditional hydrodynamics. Furthermore, the proposed sequence-to-sequence GeoPINS predicts spatially varying water depth (especially high water depth) at a much higher level of accuracy.
\subsection{Validation of Zero-shot Super-resolution}
A high-resolution numerical grid possesses the capability to accurately simulate intricate flooding behaviors. However, the execution of these models entails substantial computational expenses when adhering to traditional hydrodynamic methodologies~\citep{fraehr2023supercharging}.  In contrast, our sequence-to-sequence GeoPINS demonstrates resolution invariance, with the model being trained on a coarse grid and directly applied to a refined spatiotemporal grid for inference.
The quantitative evaluation results of this zero-shot super-resolution methodology are meticulously presented in Table~\ref{Tab:Table6}.

For spatial downscaling, we have adhered to the benchmark dataset configurations, enabling the direct utilization of the sequence-to-sequence GeoPINS trained on a coarse grid with a spatial resolution of 480m for inference at a spatial resolution of 360m.
The entire prediction time is 1451.78 seconds on an NVIDIA A6000 GPU. To gauge the model's performance, we have utilized 14-day flood depth data on the spatially refined grid, calculating daily average MAPE and MAE under various water depth thresholds (0m, 0.1m, 0.5m).  Remarkably, when the water depth threshold is set at 0m, the average MAPE over the course of 14 days stands at 22.24\%, with an average MAE of 0.0873m. Furthermore, a significant observation arises from the scrutiny of daily MAPE and MAE values over the successive 14-day period, revealing a progression in error magnitude attributed to the accumulation of initial value discrepancies, most notably pronounced on August 30 and August 31. The mitigation of these augmented errors can be achieved through the integration of real-time observations. 
Despite these challenges, it is noteworthy that the model's overall performance at a high spatial resolution remains commendable. As exemplified by the average MAPE and MAE values on August 22, which are 0.75\% and 0.0037m, respectively. Moreover, the decreasing MAE value with the increase in the water depth threshold underscores the effectiveness of sequence-to-sequence GeoPINS in forecasting high flood depths at a refined spatial resolution.

For the time downscaling experiment, we have employed the model trained on the coarse grid, characterized by a temporal resolution of 300s, for direct prediction on a finer time grid boasting a temporal resolution of 30s. Specifically, the predictions at a 300-second temporal resolution serve as the initial values for sequence-to-sequence GeoPINS on the finer time grid (30s), with predictions being executed every 10 time steps as a sequential sequence. The resulting performance metrics are meticulously outlined in Table~\ref{Tab:Table6}. Under the condition of a 0-meter water depth threshold, the average MAPE for the 14-day flood depth is determined to be 30.81\%, accompanied by an average MAE of 0.1382m. It is worth noting that this performance lags behind that achieved on the fine spatial grid. The cumulative prediction time on an NVIDIA A6000 GPU amounts to 1544.78 seconds. Again, a comparative analysis of daily MAE and MAPE values at various time points underscores the trend of increasing errors as a consequence of cumulative initial value discrepancies in sequence-to-sequence GeoPINS. Additionally, the model demonstrates a lower MAE prediction error for high water level depths at high time resolutions. 

Overall, the real-time and accuracy performance demonstrated by sequence-to-sequence GeoPINS in zero-shot super-resolution experiments at fine spatiotemporal resolutions imbue it with the potential for large-scale flood prediction.
\begin{table*}[htp!]
	\caption{Accuracy assessment of zero-shot super-resolution. }
	\centering
	\resizebox{1.0\textwidth}{!}
	{\begin{tabular}{cl|c|cccccccccccccc|c}
\hline
\multicolumn{2}{c|}{\textbf{Experiments}}                                                                            & \textbf{Forecasting Errors}                              & \textbf{18 Aug} & \textbf{19 Aug} & \textbf{20 Aug} & \textbf{21 Aug} & \textbf{22 Aug} & \textbf{23 Aug} & \textbf{24 Aug} & \textbf{25 Aug} & \textbf{26 Aug} & \textbf{27 Aug} & \textbf{28 Aug} & \textbf{29 Aug} & \textbf{30 Aug} & \textbf{31 Aug} & \textbf{Overall Errors} \\ \hline
\multicolumn{2}{c|}{\multirow{4}{*}{\makecell{Spatial   downscaling \\ (Spatial resolution=360m, \\ time resolution=300s)}}} & \makecell{MAPE (\%) with water pixels \\ (water depth $\geq$ 0m)}        & 5.95\%                  & 1.33\%                  & 0.84\%                  & 0.90\%                  & 0.75\%                  & 0.83\%                  & 0.98\%                  & 9.82\%                  & 19.72\%                 & 26.91\%                  & 45.78\%                  & 72.43\%                  & 93.64\%                  & 95.72\%                  & 22.24\%                   \\
\multicolumn{2}{c|}{}                                                                                       & \makecell{MAE (m) with water pixels \\  (water depth $\geq$ 0m) }  & 0.0101                  & 0.0038                  & 0.0032                  & 0.0037                  & 0.0037                  & 0.0044                  & 0.0053                  & 0.0527                  & 0.1015                  & 0.1151                   & 0.1598                   & 0.2420                   & 0.3872                   & 0.4290                   & 0.0873                    \\
\multicolumn{2}{c|}{}                                                                                       & \makecell{MAE (m) with water pixels  \\ (water depth $\geq$ 0.1m)} & 0.0077                  & 0.0022                  & 0.0017                  & 0.0024                  & 0.0023                  & 0.0027                  & 0.0037                  & 0.0502                  & 0.0999                  & 0.1136                   & 0.1581                   & 0.2398                   & 0.3861                   & 0.4281                   & 0.0856                    \\
\multicolumn{2}{c|}{}                                                                                       & \makecell{MAE (m) with  water pixels \\ (water depth $\geq$ 0.5m) } & 0.0061                  & 0.0015                  & 0.0011                  & 0.0017                  & 0.0016                  & 0.0021                  & 0.0029                  & 0.0364                  & 0.0732                  & 0.0865                   & 0.1331                   & 0.2081                   & 0.3467                   & 0.3946                   & 0.0724                    \\ \hline
\multicolumn{2}{c|}{\multirow{4}{*}{\makecell{Temporal   downscaling \\ (Spatial resolution=480m, \\ time resolution=30s)}}} & \makecell{MAPE (\%) with water pixels \\ (water depth $\geq$ 0m) }       & 6.17\%                  & 1.77\%                  & 1.33\%                  & 1.27\%                  & 1.75\%                  & 2.90\%                  & 4.51\%                  & 23.33\%                 & 43.57\%                 & 51.97\%                  & 72.20\%                  & 89.65\%                  & 95.31\%                  & 96.60\%                  & 30.81\%                   \\
\multicolumn{2}{c|}{}                                                                                       & \makecell{MAE (m) with water pixels   \\ (water depth $\geq$ 0m)}   & 0.0122                  & 0.0071                  & 0.0063                  & 0.0060                  & 0.0084                  & 0.0139                  & 0.0237                  & 0.1352                  & 0.2530                  & 0.2480                   & 0.2900                   & 0.3492                   & 0.4214                   & 0.4474                   & 0.1382                    \\
\multicolumn{2}{c|}{}                                                                                       & \makecell{MAE (m) with water pixels  \\ (water depth $\geq$ 0.1m)} & 0.0083                  & 0.0036                  & 0.0032                  & 0.0035                  & 0.0051                  & 0.0090                  & 0.0189                  & 0.1337                  & 0.2535                  & 0.2482                   & 0.2901                   & 0.3491                   & 0.4209                   & 0.4468                   & 0.1361                    \\
\multicolumn{2}{c|}{}                                                                                       & \makecell{ MAE (m) with  water pixels \\ (water depth $\geq$ 0.5m)}  & 0.0063                  & 0.0019                  & 0.0019                  & 0.0020                  & 0.0032                  & 0.0059                  & 0.0118                  & 0.1059                  & 0.2212                  & 0.2027                   & 0.2350                   & 0.2912                   & 0.3818                   & 0.4150                   & 0.1148                    \\ \hline
\end{tabular}}
\label{Tab:Table6}
\end{table*}

\section{Conclusion}
Our work addresses the limitations (such as model variable acquisition in data-scarce areas and issues of scalability) of existing large-scale hydrodynamic models by introducing FloodCast, a comprehensive flood modeling and forecasting framework. FloodCast combines the power of multi-satellite observations and the innovative GeoPINS hydrodynamic model. 
In the multi-satellite observation module, we present real-time UCD techniques and a rainfall processing tool, effectively utilizing multi-satellite data for large-scale flood predictions. In the hydrodynamic modeling module, a generic solver, GeoPINS, is proposed with excellent resolution-invariant (zero-shot super-resolution), geometry-adaptive properties. GeoPINS excels in performance for solving PDEs across regular and irregular domains.
Furthermore, our work extends the capabilities of GeoPINS with a sequence-to-sequence model, enabling the handling of long-term temporal series and extensive spatial domains in large-scale flood modeling. To evaluate our approach, we establish a benchmark dataset using the 2022 Pakistan flood to assess various flood prediction methods. Our validation, performed across flood inundation range, depth, and spatiotemporal downscaling transferability, employs SAR-based flood data, traditional benchmarks, and optical remote sensing images. The results reveal that traditional hydrodynamics and sequence-to-sequence GeoPINS exhibit remarkable agreement, particularly during high water levels. Moreover, comparative assessments against SAR-based flood depth data demonstrate that sequence-to-sequence GeoPINS outperforms traditional hydrodynamics with smaller prediction errors.
In summary, our method showcases the ability to maintain high-precision large-scale flood dynamics solutions. The effectiveness of zero-shot super-resolution experiments holds the potential for real-time flood prediction on high spatiotemporal resolution grids. 

One of the shortcomings in our FloodCast lies in the accumulation of initial value errors that occur during sequence-to-sequence learning.  This issue can be mitigated by providing a small amount of observation data in our work. However, it's worth noting that such data may not always be readily available in certain regions. To address this challenge in the future, we aim to further enhance FloodCast by exploring innovative approaches that combine the strengths of both ML and physical models. Our goal is to develop a fully unsupervised flood dynamics model that doesn't rely on external observation data. This represents an exciting avenue for future research, where we can continue to advance the capabilities of flood modeling and forecasting, ultimately providing more robust solutions for different regions with limited data resources.


\section*{Declaration of  Competing Interests} The authors declare that they have no competing financial interests.

\section*{Acknowledgment}
The work is jointly supported by the German Federal Ministry of Education and Research (BMBF) in the framework of the international future AI lab "AI4EO -- Artificial Intelligence for Earth Observation: Reasoning, Uncertainties, Ethics and Beyond" (grant number: 01DD20001) and the project Inno$\_$MAUS (grant number: 02WEE1632B), by German Federal Ministry for Economic Affairs and Climate Action in the framework of the "national center of excellence ML4Earth" (grant number: 50EE2201C), by the Excellence Strategy of the Federal Government and the Länder through the TUM Innovation Network EarthCare and by Munich Center for Machine Learning.


 \bibliographystyle{elsarticle-harv} 
 \bibliography{egbib}


\end{document}